%% file: main.tex
\pdfoutput=1
\documentclass[10pt,twocolumn,letterpaper]{article}
\usepackage[pagenumbers]{cvpr}
\input{preamble}
\definecolor{cvprblue}{rgb}{0.21,0.49,0.74}
\usepackage[pagebackref,breaklinks,colorlinks,allcolors=cvprblue]{hyperref}
\hypersetup{pdftitle={DriveOddsBench: Are Driving Video World Models Aligned with Real-World Odds?},pdfauthor={Jiaxi Liu, Keshu Wu, Bin Ran, Xiaopeng Li}}
\title{DriveOddsBench: Are Driving Video World Models\\Aligned with Real-World Odds?}
\author{Jiaxi Liu \qquad Keshu Wu \qquad Bin Ran \qquad Xiaopeng Li\thanks{Corresponding author.}\\
Department of Civil and Environmental Engineering, University of Wisconsin--Madison\\
{\tt\small \{jliu2487, kwu84, xli2485\}@wisc.edu, bran@engr.wisc.edu}\\
{\small Project page: \url{https://hhjj233.github.io/DriveOddsBench-page/}}\\
{\small Code: \url{https://github.com/hhjj233/DriveOddsBench}}}
\begin{document}
\maketitle
\input{sections/teaser}

\input{sections/abstract}
\input{sections/introduction}
\input{sections/related_work}
\input{sections/task}
\input{sections/theory}

\input{sections/protocol}
\input{sections/experiments}
\input{sections/discussion}
\input{sections/llm_statement}
{\small
\bibliographystyle{ieeenat_fullname}
\bibliography{bibliography/references_verified}
}
\clearpage
\appendix
\renewcommand{\thefigure}{S\arabic{figure}}\renewcommand{\thetable}{S\arabic{table}}\renewcommand{\theequation}{S\arabic{equation}}
\setcounter{figure}{0}\setcounter{table}{0}\setcounter{equation}{0}
\twocolumn[\begin{center}{\Large\bfseries Appendix}\end{center}\vspace{0.5em}]
\begin{strip}
\centering
\includegraphics[width=\linewidth]{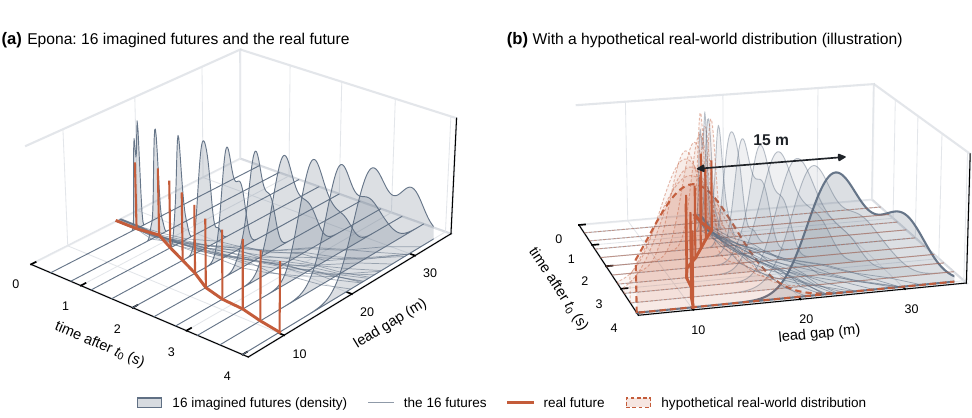}
\captionsetup{hypcap=false}
\captionof*{figure}{\textbf{The odds of a world model and the real future, over time.} One TEST history; the reading is the lead gap. \textbf{(a)}~Grey: the density of the 16 futures that Epona imagines at each time step, scaled to the same peak height; thin lines: the 16 futures. Red: the real future, read after the model's autoencoder as in the benchmark. The lead vehicle slows down to 10.2\,m at 4\,s, while all 16 futures place it 20.7--31.9\,m ahead. \textbf{(b)}~The same with a hypothetical real-world distribution (dashed), centred on the real future with the model's own spread. It is an illustration only: real-world odds are never observed, which is why the benchmark pools the ranks of real outcomes over histories. By 4\,s the median of the 16 futures lies 15\,m beyond the real future. The history was chosen to show the failure.}
\captionsetup{hypcap=true}
\end{strip}
\noindent The appendix has six parts: the theory and proofs (\cref{app:theory}), implementation and reporting details with the generation settings of every model (\cref{app:implementation}, \cref{tab:settings}), the per-model numbers behind the findings (\cref{app:details}), the whole-trajectory results (\cref{app:trajectory}), the property scores (\cref{app:scores}) and the validation of the rank test on simulated numbers (\cref{app:scalar}). The code and three short videos are on the project page: a one-minute motivating example, the pipeline, and one test history worked through.
\input{appendix/theory_proofs}
\input{appendix/implementation}
\input{appendix/detailed_results}
\input{appendix/trajectory_results}
\input{appendix/property_scores}
\input{appendix/scalar_validation}
\end{document}

%% file: preamble.tex
\usepackage[T1]{fontenc}
\usepackage{graphicx}
\usepackage{amsmath,amssymb,amsthm}
\usepackage{booktabs,array,tabularx}
\usepackage{cuted}
\usepackage{caption}
\usepackage{xcolor}
\usepackage{colortbl}
\newtheorem{theorem}{Theorem}
\newtheorem{proposition}[theorem]{Proposition}

\theoremstyle{definition}

\theoremstyle{remark}

\newcommand{\method}{DriveOddsBench}

\newcolumntype{L}[1]{>{\raggedright\arraybackslash}p{#1}}

%% file: sections/teaser.tex
\begin{strip}
\centering
\includegraphics[width=\linewidth]{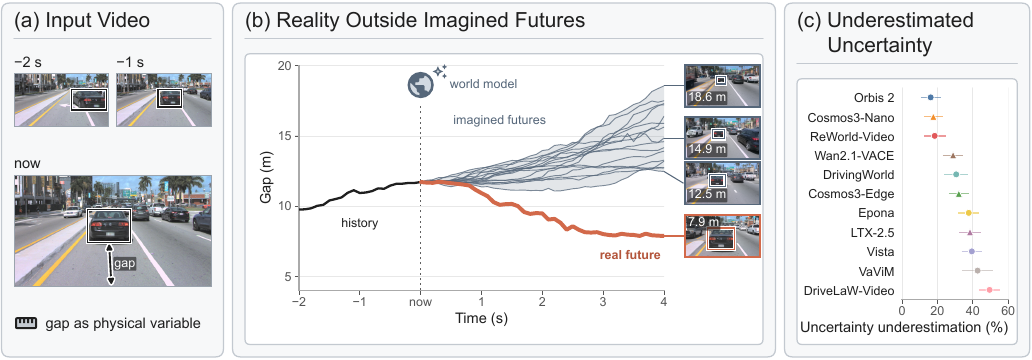}
\captionsetup{hypcap=false}
\captionof{figure}{\textbf{Driving video world models underestimate their uncertainty.} \textbf{(a)}~A test history from Argoverse~2, with the lead vehicle boxed. \textbf{(b)}~The lead gap in the real video (black, then red) and in the futures that Orbis~2 imagines (grey). The lead vehicle braked, and no imagined future did. At 4\,s the real gap lies below every imagined gap, which happens on either side in only 11.8\% of cases under the real distribution. \textbf{(c)}~Uncertainty underestimation over 200 test histories, the share of cases whose real outcome falls outside all samples minus the 11.8\% expected under the real distribution (95\% intervals). The case was chosen to show the failure.}
\label{fig:motivation}
\captionsetup{hypcap=true}
\end{strip}

%% file: sections/abstract.tex
\begin{abstract}
Video world models predict how a scene will evolve, and their sampled futures are increasingly used in place of the real one, from planning to simulation. These uses assume that the samples are aligned with real-world odds, with each future appearing among them as often as it would follow the same history in reality, yet current evaluations only score how real each sample looks. Testing this assumption on real driving data is hard for three reasons. Each recorded history has only one real future, so the true probabilities are unknown. Physical quantities must be read from video, where reading errors can look like model errors. Finally, the many possible analyses invite selective reporting. We introduce \method{}, a pre-registered benchmark that addresses all three. First, if the samples follow the real distribution, the rank of the real outcome among them is uniform in every history, so pooled ranks test them without probability labels. Second, one validated instrument reads the same quantities, such as the lead-vehicle gap and the ego motion, from the generated futures and from the real future, which first passes through the world model's autoencoder so that both are read alike. Third, all confirmatory analyses were frozen before any test clip was measured. We test eleven general and driving world models on 200 held-out Argoverse~2 histories in five scenario families, with 35{,}200 generated futures. None is aligned, and every model underestimates its uncertainty. The real outcome falls outside the range of all 16 samples in 28--61\% of cases instead of the 11.8\% expected under the real distribution, and a descriptive spread ratio that does not depend on the number of samples shows that the samples lie closer to one another than to reality. Five models are significantly skewed, in opposite directions, and the largest errors are in the ego motion. Nine models beat climatology without being aligned, and among these models neither type nor scale predicts which comes closest. We will release the instrument, code, frozen protocols and samples.
\end{abstract}

%% file: sections/introduction.tex
\section{Introduction}
\label{sec:intro}
\begin{figure*}[t]
\centering
\includegraphics[width=\linewidth]{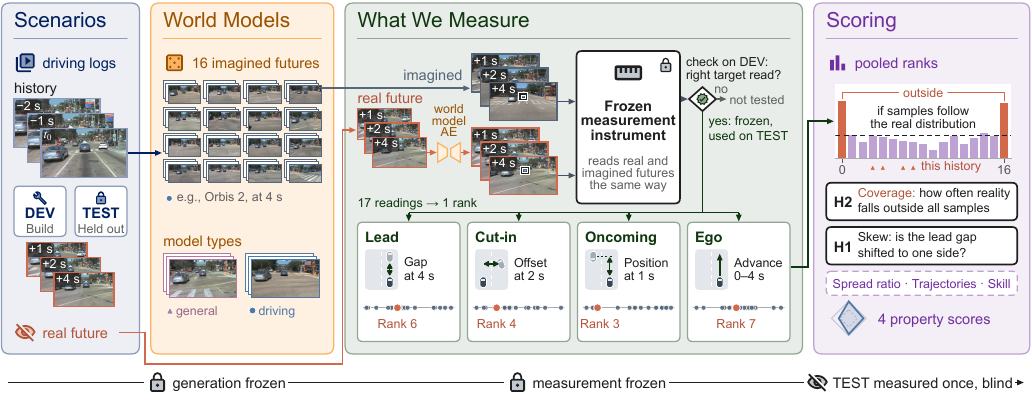}
\caption{\textbf{Overview of \method{}.} Each model continues held-out Argoverse~2 histories with 16 futures. A single frozen instrument, validated on the development split (DEV), reads the same quantities from each imagined future and from the real future after passing it through the world model's autoencoder. Pooling the rank of the real reading among the 16 samples across histories yields two tests: H2 (coverage) checks how often reality falls outside all samples, and H1 (skew) checks whether the lead gap is shifted to one side. Frame stacks show $+1$, $+2$ and $+4$\,s.}
\label{fig:method}
\end{figure*}

Video world models predict how a scene will evolve~\cite{wang2024drivedreamer,hassan2025gem,zhang2025epona,agarwal2026cosmos3}, and their sampled futures are increasingly used in place of the real one. Planners imagine the outcomes of candidate actions~\cite{wang2024drivewm,gao2024vista}, simulators synthesize rare scenarios~\cite{russell2025gaia2}, and closed-loop platforms test driving policies~\cite{yang2025drivearena}.
Each of these uses reads probabilities off the samples, so the samples should follow the real distribution, with each outcome appearing as often as it follows the same history in reality~\cite{liu2026corner}. We call such a model aligned with real-world odds.
In testing, for example, a failure rate measured in a world model predicts the real failure rate only if the world model generates each future at its real rate~\cite{feng2021nade,feng2023d2rl,montali2023wosac}.
\Cref{fig:motivation} shows how this fails: the lead vehicle braked, but it brakes in none of the 16 futures that Orbis~2 generates from the same history.

Existing benchmarks do not test this alignment.
Quality, physics and instruction-following benchmarks score one video at a time~\cite{duan2025worldscore,li2025worldmodelbench}, and feature distances compare pools of videos without a reference for each situation~\cite{unterthiner2018fvd}, so neither checks how often each outcome follows a history.
Benchmarks that do check this rely on scenes whose outcome probabilities are fixed by construction~\cite{pu2026pawbench,sadeghi2026calibench}, and driving logs offer no such scenes, because each recorded history happened only once.

Yet one real future per history is enough once many histories are pooled.
If a model's conditional distribution of a measured quantity is correct, the real outcome and the $K$ samples are exchangeable, so the rank of the real outcome among the samples is uniform on $\{0,\dots,K\}$ in every history, regardless of its outcome distribution.
Histories with different and unknown distributions therefore share one reference (\cref{fig:method}), as in ensemble weather verification~\cite{talagrand1997evaluation,hamill2001rank} and simulation-based calibration~\cite{talts2018sbc}.

\method{} turns this principle into a benchmark for driving video.
It uses 200 test histories from Argoverse~2~\cite{wilson2021argoverse2}, each with 2\,s of front-camera video and the 6\,s that followed, grouped into scenario families from information up to the last history frame.
One instrument, checked on separate development data, reads the lead gap, adjacent and oncoming vehicles and the ego motion from 16 generated futures per model and from the real future, which first passes through the world model's autoencoder so that both carry the same artefacts.
Hypotheses, instrument and confirmatory analysis were frozen before any test clip was measured.

We evaluate eleven general and driving world models, and none is aligned.
\textbf{All eleven underestimate their uncertainty}: the real outcome falls outside all 16 samples in 28--61\% of cases instead of 11.8\%, and their samples lie closer to one another than to reality.
\textbf{This comes from samples that are too narrow, shifted or both}, and five models are significantly skewed in the lead gap, in opposite directions.
\textbf{Informative samples can still miss the real distribution}, as nine models beat climatology without following it, and \textbf{neither type nor scale decides how close a model comes}.

\noindent Our contributions are:
\begin{itemize}
\item \textbf{The first test of whether driving video world models are aligned with real-world odds}, on natural driving logs where each history has only one recorded future, with scenario families defined from past information only.
\item \textbf{A protocol that makes classical rank verification valid for generated video}: one instrument that reads real and generated futures the same way, random tie-breaking, symmetric retention, pre-registered tests, and extensions to trajectories and informativeness.
\item \textbf{Pre-registered results for eleven models}, with positive controls, split-sample replication and sensitivity analyses.
\end{itemize}

%% file: sections/related_work.tex
\section{Related Work}
\label{sec:related_work}

\begin{table}[t]
\centering
\footnotesize
\setlength{\tabcolsep}{2.5pt}
\caption{\textbf{Comparison with related benchmarks.} WM: built for world models. Dist.: scores samples per input as a distribution. Odds: tests whether outcomes occur as often as samples imply, using probabilities known by construction (known) or one recorded future per history (1 future). Real: recorded future as reference.}
\label{tab:related}
\begin{tabular}{@{}l c l c c c c@{}}
\toprule
Benchmark & WM & Modality & Driving & Dist. & Odds & Real \\
\midrule
WOSAC~\cite{montali2023wosac} & $\times$ & trajectories & \checkmark & \checkmark & $\times$ & \checkmark \\
WorldScore~\cite{duan2025worldscore} & \checkmark & video & $\times$ & $\times$ & $\times$ & $\times$ \\
WorldModelBench~\cite{li2025worldmodelbench} & \checkmark & video & partly & $\times$ & $\times$ & $\times$ \\
DrivingGen~\cite{zhou2026drivinggen} & \checkmark & video & \checkmark & $\times$ & $\times$ & $\times$ \\
WorldLens~\cite{liang2026worldlens} & \checkmark & video & \checkmark & $\times$ & $\times$ & $\times$ \\
PAWBench~\cite{pu2026pawbench} & \checkmark & video & $\times$ & \checkmark & known & $\times$ \\
CaliBench~\cite{sadeghi2026calibench} & \checkmark & video & $\times$ & \checkmark & known & $\times$ \\
\midrule
\method{} (ours) & \checkmark & video & \checkmark & \checkmark & 1 future & \checkmark \\
\bottomrule
\end{tabular}
\end{table}

\paragraph{Driving world models.}
Driving world models generate future road scenes and add planning, ego-motion or scene control~\cite{wang2024drivedreamer,wang2024drivewm,gao2024vista,hassan2025gem,zhang2025epona}. General video models are also used as world simulators~\cite{agarwal2026cosmos3,hacohen2026ltx2,wan2025wan}.
\method{} treats both kinds' samples as forecasts and proposes no generator.

\paragraph{Evaluating generated videos.}
WorldScore and WorldModelBench score quality, controllability and physical adherence~\cite{duan2025worldscore,li2025worldmodelbench}, and DrivingGen and WorldLens do the same for driving~\cite{zhou2026drivinggen,liang2026worldlens}.
They score each video on its own, and FVD compares whole collections of videos in a feature space~\cite{unterthiner2018fvd}, so none checks how often each outcome occurs.

\paragraph{Probability-aware evaluation.}
PAWBench, which measures probabilistic alignment, and CaliBench compare outcome frequencies with probabilities fixed by symmetry, enumeration or physics~\cite{pu2026pawbench,sadeghi2026calibench}. We adopt CaliBench's separation of scorability from distributional error.
The Waymo Open Sim Agents Challenge scores rollouts against the logged future with estimated feature likelihoods~\cite{montali2023wosac}.
\method{} tests the odds from one recorded future per history without density estimates (\cref{tab:related}).

\paragraph{Forecast verification.}
Probability integral transforms and rank histograms compare observations against forecast distributions or ensembles~\cite{diebold1998density,gneiting2007calibration,talagrand1997evaluation,hamill2001rank}, with randomized tie-breaking for discrete outcomes~\cite{brockwell2007residuals} and a decomposition into spread and skew~\cite{jolliffe2008rank}. Simulation-based calibration uses the same ranks~\cite{talts2018sbc,sailynoja2022uniformity}, and proper scores such as the fair CRPS measure accuracy and sharpness~\cite{gneiting2007proper,ferro2014fair}.
The same tools verify generative weather forecasts against one realized future~\cite{ravuri2021dgmr,price2025gencast,rasp2024weatherbench2}, and we bring them to video world models, whose outcomes are read from pixels.

%% file: sections/task.tex
\section{\method{}}
\label{sec:task}

\subsection{Task}
An evaluation bundle for model $m$ and history $i$ is
\begin{equation}
 \mathcal B_i^{(m)}=\bigl(H_i,V_{i0},V^{(m)}_{i1},\ldots,V^{(m)}_{iK}\bigr),
 \qquad K=16,
 \label{eq:bundle}
\end{equation}
where $V_{i0}$ is the recorded future and $V^{(m)}_{ik}$ are samples drawn from the model given the conditioning $H_i$.
Every model receives the 2\,s of front-camera video before $t_0$ (21 frames at 10\,Hz), or the part of it that it accepts, and generates the next 6\,s. Epona and DrivingWorld also receive the past ego poses.
A model that reads less of the history is tested on the same histories, which are selected with information it may not see.
Models run with their official settings and a fixed seed per sample, and no sample is discarded or drawn again (details in \cref{app:implementation}).

\subsection{Histories and Scenario Families}
All histories come from the Argoverse~2 Sensor dataset~\cite{wilson2021argoverse2}.
A single dataset keeps the camera, image format and frame rate fixed, so differences between datasets cannot distort the comparison.
Anchors $t_0$ lie on a 0.5\,s grid. The history is the 2\,s before $t_0$, and the real future is the 6\,s after it.
Rules that read only annotations and poses up to $t_0$ assign each history to one or more of the five scenario families of \cref{tab:families}.
The development split (DEV) has 60 histories from 60 logs, and the test split (TEST) has 200 histories from 200 other logs.

\begin{table}[t]
\centering
\footnotesize
\setlength{\tabcolsep}{2.5pt}
\caption{\textbf{Scenario families.} Check (DEV, raw real video, against LiDAR): share of readings on the correct target and median absolute error (for Ego, odometry error on the 4\,s advance).}
\label{tab:families}
\begin{tabular}{@{}l l l r r@{}}
\toprule
Family & Target at $t_0$ & Quantity & TEST & Check \\
\midrule
Lead & moving lead & gap at 4\,s & 65 & 95\%, 1.3\,m \\
Brake/stop & braking/stopped lead & gap at 4\,s & 65 & 97\%, 1.3\,m \\
Cut-in & adjacent, same way & $|$lateral$|$ at 2\,s & 59 & 98\%, 0.2\,m \\
Oncoming & opposite way & position at 1\,s & 69 & 100\%, 1.4\,m \\
Ego & ego at junction & advance 0--4\,s & 101 & 0.3\,m \\
\bottomrule
\end{tabular}
\end{table}

\subsection{Instrument}
A readout $g$ maps a video to a scalar: $Y_i=g(H_i,V_{i0})$ and $X^{(m)}_{ik}=g(H_i,V^{(m)}_{ik})$.
It tracks the target of each family from $t_0$ with three trackers on one 512$\times$288 band of the front camera and reads distance and lateral offset from where the target's box meets the ground. Visual odometry on the same band gives the ego motion.
The same code reads every video. A lost target keeps its last reading, which a sensitivity check treats as missing (\cref{sec:checks}).
The real future first passes through the world model's autoencoder (\cref{sec:theory}), and the raw real video serves as a sensitivity check.

\paragraph{Validation and reliability labels.}
On DEV, the instrument agrees with LiDAR annotations and AV2 ego poses (\cref{tab:families}).
Before measuring any test clip of a model, we used DEV to fix which of its quantities are \emph{tested}. The correct target must be read in at least 90\% of histories on both the raw and the reconstructed real video, and the two readings must agree (Spearman $\geq0.8$).
Fourteen secondary quantities, such as headway and post-encroachment time, receive the same labels.

%% file: sections/theory.tex
\section{Testing Probabilities with Ranks}
\label{sec:theory}

\begin{table*}[t]
\centering
\caption{\textbf{Test results by property} (200 histories, $K{=}16$, sorted by H2, metrics in \cref{sec:metrics}). Brackets give 95\% intervals. \checkmark: H1 confirmed. Within each column, the best value is shaded red, second orange and third blue, with ties sharing a colour.}
\label{tab:main}
\input{tables/tab_main_results_v07.tex}
\end{table*}

\begin{figure*}[t]
\centering
\includegraphics[width=\linewidth]{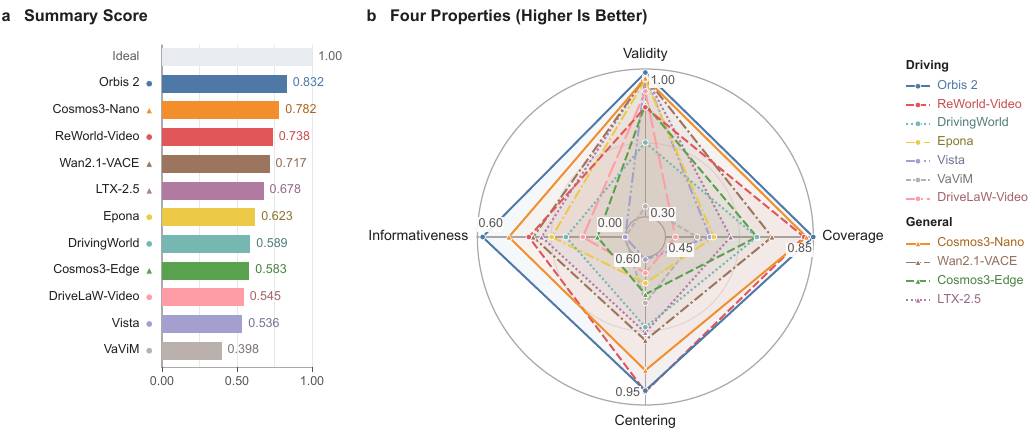}
\caption{\textbf{Property scores on TEST} (descriptive, 1 is ideal). \textbf{(a)}~Summary score (mean of the four properties). \textbf{(b)}~Individual property scores. Each axis spans the observed model range, so radii are not comparable across axes. Marker: model type ($\bullet$ driving, $\blacktriangle$ general).}
\label{fig:overview}
\end{figure*}

\begin{figure}[t]
\centering
\includegraphics[width=\linewidth]{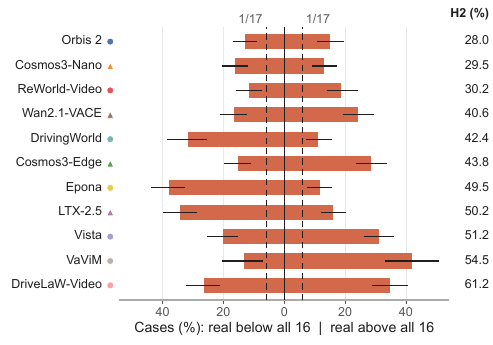}
\caption{\textbf{Every model underestimates its uncertainty.} Share of cases with the real outcome below (left) or above (right) all 16 samples, with 95\% intervals. Dashed: 1/17 per side, the value under the real distribution. Right: H2.}
\label{fig:coverage}
\end{figure}

\begin{figure*}[t]
\centering
\includegraphics[width=\linewidth]{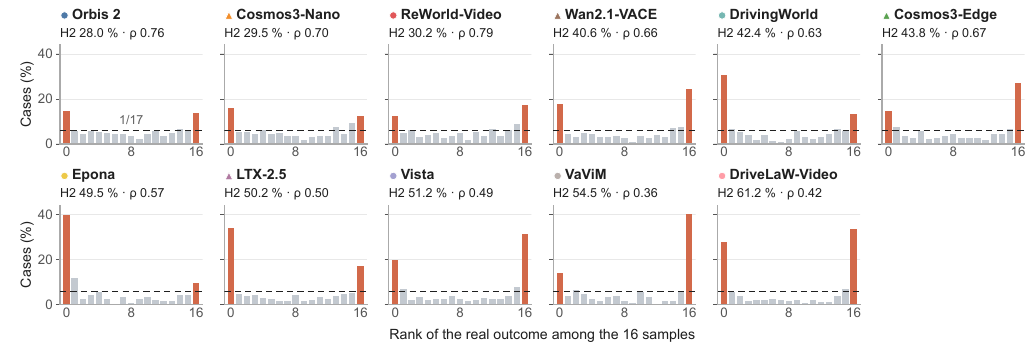}
\caption{\textbf{Rank histograms} (TEST), pooled over the primary quantities of the scored families, one panel per model in the order of \cref{tab:main}. Dashed: 1/17, the value under the real distribution. Terracotta: ranks 0 and 16, where the real outcome lies outside all samples. Labels: H2 and the spread ratio $\rho$, which is 1 under the real distribution.}
\label{fig:ranks}
\end{figure*}

\begin{figure*}[t]
\centering
\includegraphics[width=\linewidth]{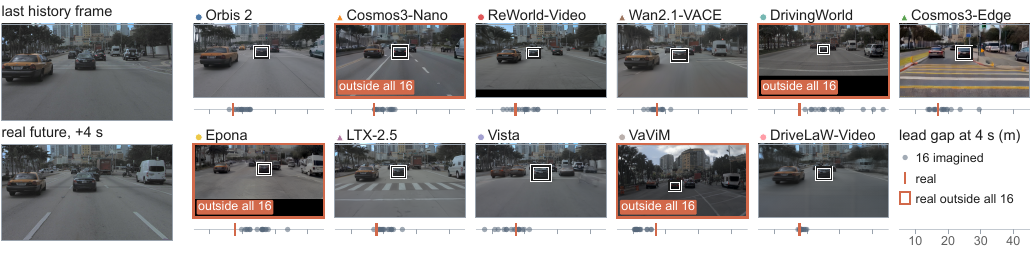}
\caption{\textbf{One typical test history, eleven models.} Left: the last history frame and the real future at $+4$\,s. Each model: its median imagined future at $+4$\,s with the instrument's box, and below it the 16 imagined lead gaps (slate) and the real gap (terracotta). Framed: the real gap lies outside all 16. Chosen by rule as typical: four models miss here, the median over the 57 histories that every model scores.}
\label{fig:gallery}
\end{figure*}

\begin{figure*}[t]
\centering
\includegraphics[width=\linewidth]{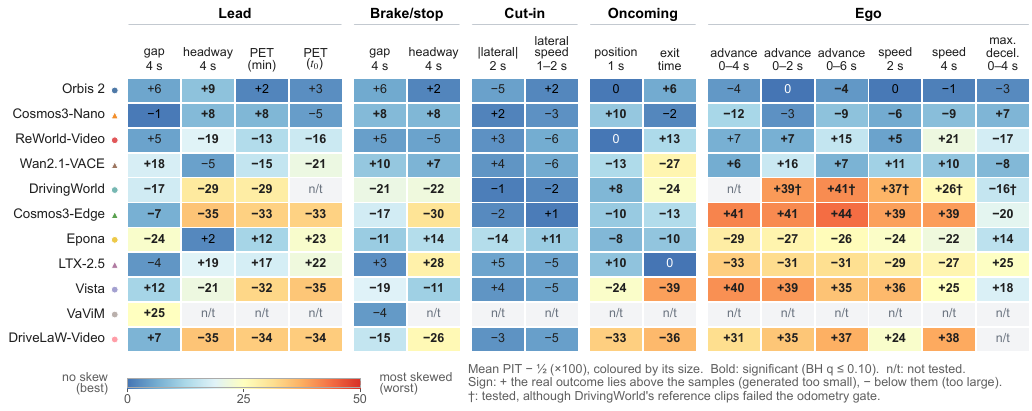}
\caption{\textbf{Ego motion shows the largest skew.} Mean $\mathrm{PIT}-\tfrac12$ ($\times100$) on TEST for every quantity tested for more than two models. Colour shows offset size, from blue (none) to red (50 or more). Sign: $+$ means generated values are too small, and $-$ means they are too large. Bold: significant (BH $q\le0.10$). n/t: not tested. PET: post-encroachment time. $\dagger$: tested, although DrivingWorld's reference clips failed the odometry gate.}
\label{fig:profile}
\end{figure*}

\subsection{A Shared Reference without Probability Labels}
Fix a readout and a horizon, and let $P_H$ and $Q_H$ be the conditional laws of the real and generated readings given the conditioning $H$. The null hypothesis is that the model is aligned with real-world odds, that is, the samples follow the real distribution, which the forecasting literature calls calibration:
\begin{equation}
 \mathcal H_0:\quad P_H=Q_H\quad\text{for almost every }H.
 \label{eq:semantic-null}
\end{equation}
Given $H_i$, the $K$ generated readings are independent draws from $Q_{H_i}$ and are independent of $Y_i$.
With independent tie-breaking tags $U_{i0},\dots,U_{iK}\sim\operatorname{Unif}(0,1)$, the rank of the real reading is
\begin{equation}
R_i=\sum_{k=1}^{K}\Bigl(\mathbf 1\{X_{ik}<Y_i\}+\mathbf 1\{X_{ik}=Y_i,\,U_{ik}<U_{i0}\}\Bigr).
\label{eq:randomized-rank}
\end{equation}

\begin{theorem}[Uniform rank reference]
\label{thm:rank-uniform}
If the real and generated readings are exchangeable given $H_i$, which holds under \eqref{eq:semantic-null} and the sampling assumption, then
$\Pr(R_i=r\mid H_i)=1/(K{+}1)$ for $r=0,\dots,K$.
The same holds conditional on retaining a bundle when retention requires all $K{+}1$ readings to pass one shared validity rule.
\end{theorem}
\begin{proof}
Write $W_{i0}=Y_i$ and $W_{ik}=X_{ik}$. Given $H_i$, the $K{+}1$ pairs $(W_{ij},U_{ij})$ are exchangeable, and the tags make them distinct almost surely. Let $A_{jr}$ indicate that pair $j$ takes position $r$ in their lexicographic order, so that $R_i=r$ exactly when $A_{0r}=1$. Each position is taken by exactly one pair, so $\sum_{j}\Pr(A_{jr}=1\mid H_i)=1$, and exchangeability makes the $K{+}1$ terms equal. Hence $\Pr(R_i=r\mid H_i)=\Pr(A_{0r}=1\mid H_i)=1/(K{+}1)$. Retention that depends only on the validity of all $K{+}1$ readings is symmetric in the labels, so the argument still holds.
\end{proof}
\noindent Because this reference does not depend on $H_i$, ranks from different histories can be pooled.
Readings can tie, for example when a lost target keeps its last value. We break ties at random, since strict comparisons would push every tie to one end and reject a correct model (\cref{app:ties}).
Reading the real future after the world model's autoencoder keeps the readings exchangeable, since real and generated clips then share the same decoder artefacts.
The theorem also implies two rules: a bundle enters a quantity only if all $K{+}1$ readings are valid, and no sample is selected for quality.

\subsection{Coverage and Skew}
With an independent uniform $U'_i$, $\mathrm{PIT}_i=(R_i+U'_i)/(K{+}1)$ is uniform on $(0,1)$ under $\mathcal H_0$.
\textbf{Coverage (H2).} Under $\mathcal H_0$, the real reading falls outside the range of all samples with probability $2/(K{+}1)=11.8\%$ for every $H$. When this happens more often, the samples are more certain about the real outcome than the model's errors allow, and we say that the model underestimates its uncertainty. Too little spread and a shift both cause this. Under $\mathcal H_0$ the real reading falls below all samples, or above them, with probability $1/(K{+}1)$ each, so an excess on one side indicates a shift, and on both sides samples that are too narrow. H2 averages the outside rate over each history's scored quantities and then over histories to form $T$, and tests $T$ against a permutation null that swaps the real video with the samples within each history.
\textbf{Spread ratio.} Unlike this outside rate, the spread ratio $\rho=\mathbb E|X-X'|/\mathbb E|X-Y|$ does not depend on $K$. It divides the mean distance between two samples by the mean distance from a sample to reality (ratio of sums per family). It equals 1 under $\mathcal H_0$. Both distances, the two terms of the fair CRPS, are unbiased for every $K\ge2$~\cite{ferro2014fair}, and values below 1 mean that the samples lie closer to each other than to reality.
\textbf{Skew (H1).} A skewed model places its samples to one side of reality, so the real outcome falls toward one end of them. Skew here is not the skewness of a distribution. We test $\overline{\mathrm{PIT}}-\tfrac12$, which has exact null variance $1/(12n)$, with the slope component of the rank chi-square~\cite{jolliffe2008rank}. When $\overline{\mathrm{PIT}}>\tfrac12$, the real outcome tends to lie above the samples, so generated values are too small.
Secondary tests and the fair CRPS are in \cref{app:secondary}.

\subsection{Whole Trajectories}
\label{sec:whole-trajectories}
A scalar at one horizon cannot tell apart continuations that reach the same value in different ways, so we also rank whole trajectories.
Any \emph{pre-rank} that maps each of the $K{+}1$ trajectories to a scalar, treating all of them alike, keeps the rank uniform~\cite{thorarinsdottir2016assessing}.
We use the band depth, which measures how central a trajectory is. At every time step the $K{+}1$ trajectories are ranked, and a trajectory with rank $r$ adds $r(K{+}2-r)$, which is largest in the middle. Under $\mathcal H_0$, the real trajectory is the least central of the $K{+}1$ with probability $1/(K{+}1)=5.9\%$. Whole trajectories are scored with the fair energy score~\cite{gneiting2007proper}.

\subsection{Real Distribution and Informativeness}
\label{sec:informativeness}
Uniform ranks check the probabilities only on average over histories, so whether the samples use the history needs a second measure. Let $\bar P$ be the climatological law of a reading, the mixture of $P_H$ over the histories of a family, and let the \emph{climatological forecast} draw its samples from $\bar P$ whatever the history.
\begin{proposition}[Climatology passes the pooled rank tests]
\label{prop:climatology}
If the histories are drawn from the population that defines $\bar P$, the rank of the real reading among $K$ draws from $\bar P$ is uniform on $\{0,\dots,K\}$, although $\bar P\neq P_H$ whenever the history is informative.
\end{proposition}
\noindent Pooled ranks therefore cannot reject a forecast that ignores the scene, but a proper score can. Averaging the decomposition of the expected CRPS over histories (\cref{app:informativeness}) gives, for a model $Q$,
\begin{equation}
 \mathrm{CRPSS}_Q=1-\frac{U+D_Q}{U+I}=\frac{I-D_Q}{U+I}\le\frac{I}{U+I},
 \label{eq:skill}
\end{equation}
where $U=\mathbb E_H\!\int P_H(1-P_H)\,dx$ is the outcome's uncertainty given the history, $I=\mathbb E_H\!\int(P_H-\bar P)^2dx$ is the information that the history carries, and $D_Q=\mathbb E_H\!\int(Q_H-P_H)^2dx$ is the distance of the model from reality. Under the CRPS, $U+I$ is the generalized entropy of the climatological law, and $I$ is the part of it that the history explains, the counterpart of mutual information~\cite{gneiting2007proper}.
We call $\mathrm{CRPSS}_Q$ the \emph{informativeness} of the model. Climatology scores 0, a model scores above 0 only if $D_Q<I$, and only $Q_H=P_H$, which is $\mathcal H_0$ itself, reaches the bound. The fair energy score has the same structure for whole trajectories.

\subsection{Four Properties and Their Scores}
\label{sec:scores}
We summarize each model by four scores in $[0,1]$, with 1 ideal. \emph{Validity} multiplies the share of usable cases by the share of clips without a break at $t_0$. \emph{Coverage} compares H2 and the least-central share with their values under $\mathcal H_0$ and penalizes too narrow and too wide samples alike. \emph{Centering} is one minus twice the mean $|\overline{\mathrm{PIT}}-\tfrac12|$, and \emph{informativeness} averages the CRPS and energy-score skills.
They summarize \cref{tab:main}, are not tests, and are defined in \cref{app:scores}.

%% file: tables/tab_main_results_v07.tex
\definecolor{rankfirst}{RGB}{248,203,203}
\definecolor{ranksecond}{RGB}{253,228,200}
\definecolor{rankthird}{RGB}{208,226,244}
\setlength{\tabcolsep}{3.6pt}
\footnotesize
\begin{tabular}{@{}l l r r l l r l r r r@{}}
\toprule
 & & \multicolumn{2}{c}{Validity} & \multicolumn{3}{c}{Coverage (H2)} & \multicolumn{2}{c}{Skew (H1)} & \multicolumn{2}{c}{Informativeness} \\
\cmidrule(lr){3-4}\cmidrule(lr){5-7}\cmidrule(lr){8-9}\cmidrule(l){10-11}
Model & Type & \begin{tabular}[b]{@{}r@{}}Scor.\\\%\,$\uparrow$\end{tabular} & \begin{tabular}[b]{@{}r@{}}Jump\\\%\,$\downarrow$\end{tabular} & \begin{tabular}[b]{@{}l@{}}Fixed time\\point (\%)\end{tabular} & Traj. (\%) & \begin{tabular}[b]{@{}r@{}}Spread\\ratio\end{tabular} & H1 & \begin{tabular}[b]{@{}r@{}}$|\mathrm{PIT}-\frac{1}{2}|$\\$\downarrow$\end{tabular} & \begin{tabular}[b]{@{}r@{}}Fixed time\\point\,$\uparrow$\end{tabular} & Traj.\,$\uparrow$ \\
\midrule
Orbis 2 & driving & \cellcolor{rankfirst}98.3 & \cellcolor{rankfirst}0.0 & \cellcolor{rankfirst}28.0\,{\scriptsize[22.5, 33.6]} & \cellcolor{rankfirst}16.9\,{\scriptsize[12.5, 21.7]} & \cellcolor{ranksecond}0.76 & $+0.056$\,{\scriptsize[$-0.027$, $+0.137$]} & \cellcolor{ranksecond}4.2 & \cellcolor{rankfirst}$+0.56$ & \multicolumn{1}{>{\columncolor{rankfirst}[\tabcolsep][0pt]}r@{}}{$+0.60$} \\
Cosmos3-Nano & general & \cellcolor{rankthird}97.8 & 2.1 & \cellcolor{ranksecond}29.5\,{\scriptsize[24.0, 35.2]} & \cellcolor{ranksecond}18.7\,{\scriptsize[13.9, 23.5]} & \cellcolor{rankthird}0.70 & \cellcolor{rankfirst}$-0.008$\,{\scriptsize[$-0.092$, $+0.077$]} & \cellcolor{rankthird}6.6 & \cellcolor{ranksecond}$+0.45$ & \multicolumn{1}{>{\columncolor{ranksecond}[\tabcolsep][0pt]}r@{}}{$+0.50$} \\
ReWorld-Video & driving & 82.2 & 0.4 & \cellcolor{rankthird}30.2\,{\scriptsize[24.4, 36.4]} & \cellcolor{rankthird}19.1\,{\scriptsize[14.0, 24.3]} & \cellcolor{rankfirst}0.79 & \cellcolor{rankthird}$+0.051$\,{\scriptsize[$-0.028$, $+0.130$]} & \cellcolor{rankfirst}4.1 & \cellcolor{rankthird}$+0.37$ & \multicolumn{1}{>{\columncolor{rankthird}[\tabcolsep][0pt]}r@{}}{$+0.41$} \\
Wan2.1-VACE & general & \cellcolor{rankfirst}98.3 & 2.2 & 40.6\,{\scriptsize[34.9, 46.4]} & 24.4\,{\scriptsize[19.7, 29.8]} & 0.66 & $+0.177$\,{\scriptsize[$+0.088$, $+0.261$]}\,\checkmark & 10.2 & $+0.34$ & \multicolumn{1}{>{\columncolor{rankthird}[\tabcolsep][0pt]}r@{}}{$+0.41$} \\
DrivingWorld$^{\ddagger\S}$ & driving & 65.2 & \cellcolor{rankfirst}0.0 & 42.4\,{\scriptsize[35.7, 49.1]} & 30.2\,{\scriptsize[23.7, 36.3]} & 0.63 & $-0.169$\,{\scriptsize[$-0.252$, $-0.087$]}\,\checkmark & 11.7 & $+0.18$ & $+0.30$ \\
Cosmos3-Edge & general & \cellcolor{rankthird}97.8 & 14.9 & 43.8\,{\scriptsize[38.2, 49.6]} & 28.6\,{\scriptsize[23.8, 33.8]} & 0.67 & $-0.066$\,{\scriptsize[$-0.148$, $+0.016$]} & 15.6 & $+0.06$ & $+0.17$ \\
Epona$^{\ddagger}$ & driving & 95.0 & \cellcolor{rankfirst}0.0 & 49.5\,{\scriptsize[43.4, 55.5]} & 44.6\,{\scriptsize[39.0, 50.0]} & 0.57 & $-0.238$\,{\scriptsize[$-0.310$, $-0.161$]}\,\checkmark & 16.9 & $+0.26$ & $+0.33$ \\
LTX-2.5 & general & 96.9 & \cellcolor{ranksecond}0.1 & 50.2\,{\scriptsize[44.0, 56.4]} & 35.3\,{\scriptsize[30.0, 41.1]} & 0.50 & \cellcolor{ranksecond}$-0.039$\,{\scriptsize[$-0.132$, $+0.054$]} & 11.1 & $+0.30$ & $+0.38$ \\
Vista$^{\dagger}$ & driving & \cellcolor{ranksecond}98.1 & 1.4 & 51.2\,{\scriptsize[45.4, 56.7]} & 44.7\,{\scriptsize[39.1, 50.2]} & 0.49 & $+0.120$\,{\scriptsize[$+0.029$, $+0.206$]}\,\checkmark & 19.7 & $-0.17$ & $-0.01$ \\
VaViM$^{*}$ & driving & 35.4 & 1.3 & 54.5\,{\scriptsize[45.5, 63.4]} & 47.8\,{\scriptsize[37.9, 57.1]} & 0.36 & $+0.249$\,{\scriptsize[$+0.165$, $+0.328$]}\,\checkmark & 14.6 & $-0.30$ & $-0.04$ \\
DriveLaW-Video & driving & 89.7 & \cellcolor{rankthird}0.3 & 61.2\,{\scriptsize[55.2, 67.3]} & 51.6\,{\scriptsize[44.8, 58.2]} & 0.42 & $+0.072$\,{\scriptsize[$-0.027$, $+0.167$]} & 18.1 & $+0.12$ & $+0.23$ \\
\bottomrule
\end{tabular}

\smallskip
\parbox{\linewidth}{\scriptsize $^{\dagger}$\,Given only the last history frame. $^{\ddagger}$\,Also receives the past ego poses. $^{\S}$\,Ego family not scored. $^{*}$\,Lead and brake/stop families only.}

%% file: sections/protocol.tex
\subsection{Pre-registered Protocol}
\label{sec:protocol}

\paragraph{Confirmatory hypotheses.}
\textbf{H2} states that the real outcome falls outside all samples too often, that is, the model underestimates its uncertainty. It is tested one-sided with $T$ over the five scored families. \textbf{H1} states that the lead gap at 4\,s is skewed. It is tested two-sided and confirmed only if the Holm-adjusted $p\le0.05$ \emph{and} the 95\% bootstrap interval of $\overline{\mathrm{PIT}}-\tfrac12$ excludes zero on the same side.
Holm's procedure controls the family-wise error over H2 and H1~\cite{holm1979simple}. Secondary tests use Benjamini--Hochberg control at $q=0.10$~\cite{benjamini1995fdr}, and everything else is exploratory.

\paragraph{Freezing and blind measurement.}
Each model's generation was frozen before its test futures were generated, and its measurement and analysis before any of its test clips was measured, with SHA-256 hashes.
No test clip was viewed before the frozen analyses ran, and every amendment is documented with the released code. Every analysis runs a positive control, a split-sample replication and pre-listed sensitivity analyses (\cref{sec:checks}).
The trajectory analysis was added after the scalar results and frozen before any trajectory statistic was computed, so it is exploratory. The spread ratio was added after all results and is descriptive.

%% file: sections/experiments.tex
\section{Experiments}
\label{sec:experiments}

\subsection{Setup}
\paragraph{Models.}
We evaluate four general world models, Cosmos3-Nano and Cosmos3-Edge~\cite{agarwal2026cosmos3}, LTX-2.5 (two-stage distilled)~\cite{hacohen2026ltx2} and Wan2.1-VACE (1.3B)~\cite{jiang2025vace}, and seven driving world models, Orbis~2~\cite{mittal2026orbis2}, Epona~\cite{zhang2025epona}, DrivingWorld~\cite{hu2026drivingworld}, ReWorld-Video~\cite{xia2026reworld}, DriveLaW-Video~\cite{xia2026drivelaw}, VaViM~\cite{bartoccioni2025vavim} and Vista~\cite{gao2024vista}. The two Cosmos3 models, which VehDyn~\cite{wang2026vehdyn} calls physical world models, count as general.
Five models read only part of the history. DrivingWorld reads the last 1.5\,s, ReWorld-Video and DriveLaW-Video the last 1.0\,s at 8\,Hz, VaViM five frames at 2\,Hz, and Vista the last frame.

\paragraph{Metrics.}
\label{sec:metrics}
\Cref{tab:main} reports the metrics of each property, and \cref{fig:overview} the scores.
Validity uses Scor., the share of usable (history, family) cases, where a family that cannot be read counts as unusable, and Jump, the share of generated clips that break at $t_0$ (at most 1.5\% for the reference clips).
Coverage uses the share of cases whose real outcome lies outside all 16 samples at a fixed time point (H2, 11.8\% under the real distribution), the share whose real trajectory is the least central of the 17 (5.9\%), and the spread ratio (1).
Skew uses H1, the mean $\mathrm{PIT}-\tfrac12$ of the lead gap, and $|\mathrm{PIT}-\tfrac12|$, the mean absolute offset over the scored primary quantities ($\times100$). Informativeness uses the CRPS and energy-score skills over climatology.
Higher is better for Scor.\ and the skills, lower for Jump and $|\mathrm{PIT}-\tfrac12|$, closer to the real-distribution value for coverage, and nearer 0 for H1.

\subsection{Main Results}
\label{sec:main-results}
Orbis~2, Cosmos3-Nano and ReWorld-Video achieve the highest summary scores, while VaViM, Vista and DriveLaW-Video score lowest (\cref{fig:overview}).

\paragraph{Most generated futures can be read.}
Seven models yield scorable readings in 95--98\% of cases (\cref{tab:main}).
ReWorld-Video and DriveLaW-Video lose cases because they overwrite history frames, while DrivingWorld and VaViM lose families that cannot be read reliably on their own reconstructions.
Cosmos3-Edge jumps at $t_0$ in 14.9\% of clips, compared with at most 2.2\% for every other model, accounting for part of its downstream error.

\paragraph{Every model underestimates its uncertainty (H2).}
H2 is confirmed for all eleven models (Holm $p\le2\times10^{-5}$, \cref{tab:main,fig:coverage}). The real outcome falls outside all 16 samples in 28--61\% of cases, 2.4 to 5.2 times the 11.8\% expected under the real distribution, and five models miss the ego advance in 70\% of histories or more.
The excess is present with two samples, grows with the number of samples, and survives every check in \cref{sec:checks}.
The spread ratio, which does not depend on $K$, agrees (\cref{fig:ranks}), at 0.36 to 0.79 with every 95\% interval below 1, and changes by at most 0.01 with 2, 4 or 8 samples (\cref{tab:spread}).

\paragraph{Five models are skewed, in opposite directions (H1).}
Epona and DrivingWorld place the lead vehicle too far ahead, and VaViM, Wan2.1-VACE and Vista place it too close, with $|\overline{\mathrm{PIT}}-\tfrac12|$ up to 0.25 (\cref{tab:main}).
In \cref{fig:gallery}, three models place the lead too far ahead and VaViM too close.

\paragraph{Informative samples still miss the real distribution.}
Nine models remove part of the error of climatology (\cref{sec:informativeness,tab:main}). Orbis~2 removes the most (56\%), yet none follows the real distribution.
Informativeness orders the models much as coverage does, as \eqref{eq:skill} predicts, because a larger distance $D_Q$ means more misses and less skill.

\subsection{Further Analyses}
\label{sec:exploratory}
\paragraph{Whole trajectories are missed even more often.}
The real trajectory is least central among the 17 in 17--52\% of cases, compared with 5.9\% under the real distribution, in nearly the same model order. For every model except Cosmos3-Nano, this excess appears within 0.4\,s of $t_0$.

\paragraph{Shift and spread are different failures.}
Orbis~2, Cosmos3-Nano and ReWorld-Video miss on both sides and have U-shaped histograms, so their samples are roughly centred but too narrow (\cref{fig:coverage,fig:ranks}). Epona, DrivingWorld and LTX-2.5 miss mostly below, and Cosmos3-Edge, Vista and VaViM mostly above, so a shift is their main error.

\paragraph{The largest skew is in the ego motion.}
Cosmos3-Nano, Epona and LTX-2.5 move the ego too far, and Vista, Cosmos3-Edge, DriveLaW-Video and DrivingWorld too little (\cref{fig:profile}).
Headway and post-encroachment time depend on ego speed, so they inherit these errors.
Five models reach their full offset within 0.8\,s, so their initial speed is wrong.

\paragraph{Model type and scale do not explain the ranking.}
Among our eleven models, driving world models rank first and in the bottom three, and general models 2 to 8 (\cref{fig:overview}).
The 1B Orbis~2 beats the 16B Cosmos3-Nano and 22B LTX-2.5, and two 2B models given the same last second (ReWorld-Video, DriveLaW-Video) rank third and ninth.

\subsection{Robustness Checks}
\label{sec:checks}
The positive control, which uses one generated sample as the observation, is close to uniform for every model ($|\mathrm{PIT}-\tfrac12|\le0.09$ per family, also over whole trajectories), so the instrument and the analysis do not create the departures.
Both halves of the samples reproduce every confirmatory decision, and neither the raw real video nor treating lost targets as missing changes a conclusion.
In simulations with known answers, the rank test rejects a correct model about 5\% of the time (\cref{app:scalar}).

%% file: sections/discussion.tex
\section{Discussion and Conclusion}
\label{sec:discussion}

\paragraph{Implications for users of world models.}
\emph{Plausible samples are not enough for a forecast.} A model that underestimates its uncertainty gives outcomes away from its samples too little probability, yet risk estimation and testing rely on these probabilities~\cite{liu2026corner}.
\emph{Report coverage with the sample budget}, as few samples hide the excess, or the spread ratio, which does not depend on it.
\emph{Spread and centre need different fixes.} Narrow models need more diverse samples, and shifted models a corrected mean, mostly of the ego speed.

\paragraph{Limitations.}
The instrument has errors (median 1.3\,m for the lead gap), but they are controlled: it reads real and generated futures alike, so errors shared by both cannot cause a rejection, and the appendix details its validation, reliability labels and sensitivity checks. Every conclusion still concerns the measured quantities.
Uniform pooled ranks would not prove alignment, because a fixed $K$ sees finitely many features of the PIT law, pooling can cancel opposite errors~\cite{hamill2001rank}, and five scalar quantities do not determine a video. A non-rejection therefore certifies nothing.
Our histories are at most 2\,s long and come from US cities, and AV2 may be closer to some models' training data. Per-family tests detect mean-PIT shifts of about 0.10.

\paragraph{Conclusion.}
A world model can render every future convincingly and still get the odds wrong. None of the eleven tested models is aligned with real-world odds: all underestimate their uncertainty, and five are significantly skewed.
Such alignment should become a standard requirement.

%% file: sections/llm_statement.tex
\section*{LLM Usage Statement}
We used ChatGPT and Codex (OpenAI), Claude (Anthropic) and Gemini (Google) to assist with writing and running code for video generation, measurement, analysis and figures, and with drafting and editing text, including the proofs in the appendix. The authors checked all code, results, proofs and text produced with these tools and take full responsibility for the content of the paper.

%% file: appendix/theory_proofs.tex
\section{Theory and proofs}
\label{app:theory}

The uniformity of ranks itself is standard in ensemble forecasting and simulation-based calibration~\cite{hamill2001rank,talts2018sbc}.

\subsection{Assumptions}
\label{app:probability-model}
$H$ is the entire conditioning tuple, including any poses or text a model receives. For one readout and horizon, $P_H$ and $Q_H$ are the laws of the measured real and generated outcomes, and the null $\mathcal H_0$ is their equality for almost every history. The null concerns the measured quantity and not the full law of the video, so the instrument has to be validated separately. With $W_{i0}=Y_i$ and $W_{ik}=X_{ik}$, \Cref{thm:rank-uniform} needs four assumptions. (i) Given $H_i$, $Y_i\sim P_{H_i}$ and $X_{i1},\ldots,X_{iK}$ are independent draws from $Q_{H_i}$, independent of $Y_i$ (it suffices that $W_{i0},\dots,W_{iK}$ are exchangeable given $H_i$ under the null). (ii) Readouts, horizons and retention rules are fixed without regard to the test outcomes. (iii) Tie tags are uniforms independent of each other and of the readings. (iv) For the across-history null, bundles are independent. A sample selected for its content breaks (i), and two histories from one log can break (iv). The protocol therefore keeps every sample and takes at most one history per log.

\subsection{Exact treatment of ties}
\label{app:ties}
With the tie-breaking tags $U_{i0},\dots,U_{iK}$ of \cref{eq:randomized-rank}, let $L_i$ count generated values strictly below $Y_i$ and $E_i$ count those equal to $Y_i$. An equivalent implementation is
\begin{equation}
 R_i=L_i+J_i,\qquad J_i\sim\operatorname{Unif}\{0,\ldots,E_i\},
 \label{eq:app-direct-ties}
\end{equation}
since the real reading's tag takes a uniform position among the $E_i+1$ tied tags. Counting only strict inequalities pushes every tied observation to the low end and causes a correct model to fail the uniformity test. We fix the tie-breaking seed in advance.

\subsection{Proof of the uniform reference}
\label{app:uniform-proof}
We restate the proof of \cref{thm:rank-uniform} for one history and add the details of retention.
Fix a history $h$ under the null. The pairs $Z_j=(W_j,U_j)$, $j=0,\ldots,K$, are exchangeable given $h$ and almost surely distinct. Let $A_{jr}$ indicate that pair $j$ takes position $r$ in the lexicographic order. Since $\sum_jA_{jr}=1$ and exchangeability makes the $\mathbb E[A_{jr}\mid h]$ equal, $\Pr(R=r\mid h)=\mathbb E[A_{0r}\mid h]=1/(K{+}1)$. No conditional CDF of the real world is estimated, since each history contributes one rank and all histories share the same reference.

\paragraph{Symmetric retention.} Suppose each reading records a validity flag alongside its value, the $Z_j$ with these flags are exchangeable given $h$, and the retention event $A$ is invariant to permutations of the $K{+}1$ labels. For every measurable set $B$ and permutation $\pi$, $\Pr(Z\in B\mid h,A)=\Pr(\pi Z\in B\mid h,A)$ by exchangeability applied to $B\cap A$ with $\pi A=A$. Retaining a bundle only when all $K{+}1$ readings are valid therefore preserves the uniform reference. Filtering on the real future alone, applying asymmetric thresholds to real and generated clips, or selecting samples by their readings breaks this symmetry.

\subsection{Across-history inference}
\label{app:inference}
Across $N$ independent bundles, the bin counts $C$ of the ranks follow a multinomial distribution with $N$ trials and bin probabilities $1/(K{+}1)$. For a pre-specified test statistic $T(C)$ and $B$ simulated null count vectors,
\begin{equation}
 p_{\mathrm{MC}}=\frac{1+\sum_{b=1}^{B}\mathbf 1\{T(C^{(b)})\ge T(C)\}}{B+1}
 \label{eq:app-mc}
\end{equation}
satisfies $\Pr(p_{\mathrm{MC}}\le\alpha)\le\alpha$ because the $B{+}1$ statistics are exchangeable under the null.

\paragraph{The null of H2.} The H2 statistic $T$ averages indicators that are dependent within a history, since one real video gives all of its readings. Its null therefore comes from a permutation test instead. Under $\mathcal H_0$ for the joint law of a history's readings, the real video is exchangeable with the generated videos of its history, so each permutation picks, for each history, one of its real and generated videos at random as the observation for all its quantities at once, which keeps their dependence. We use $B=100{,}000$ permutations and the plus-one formula \eqref{eq:app-mc}. Intervals are percentile bootstrap intervals over histories.

\subsection{What pooled ranks cannot detect}
\label{app:finite-k}
\emph{Finite resolution.} Given a continuous model CDF $F$, $R\mid F(Y)\sim\operatorname{Binomial}(K,F(Y))$. Uniform ranks at sample size $K$ match only the first $K$ moments of $F(Y)$ to those of the uniform distribution. With $L_{K+1}$ denoting the shifted Legendre polynomial of degree $K{+}1$, the density $1+\eta L_{K+1}(t)$ of $F(Y)$ for small $\eta\neq0$ yields uniform ranks at $K=16$ even though $P\neq Q$.

\emph{Pooling.} Suppose $H=\pm1$ with equal probability, $Q$ is uniform on $[0,1]$, and $p_h(y)=1+h\eta(2y-1)$. Then
\begin{equation}
 \Pr(R=r\mid H=h)=\frac1{K+1}\Bigl[1+h\eta\,\frac{2r-K}{K+2}\Bigr],
 \label{eq:app-conditional-rank-bias}
\end{equation}
so each history has a tilted rank distribution while the pooled rank distribution remains uniform~\cite{hamill2001rank}.

\emph{Projections.} Matching the marginal distributions of finitely many scalar readouts determines neither their joint distribution nor the full video distribution. Every claim is therefore indexed by quantity, horizon, conditioning and history population, and failing to reject the null does not prove that a model matches reality.

\subsection{Real distribution and informativeness}
\label{app:informativeness}
\paragraph{Climatology passes the pooled rank tests.}
If $H$ is drawn from a family's population, $Y\mid H\sim P_H$, and $X_1,\dots,X_K$ are drawn i.i.d.\ from $\bar P=\mathbb E_HP_H$, independently of $H$ and $Y$, then $\Pr(Y\in B_0,X_1\in B_1,\dots,X_K\in B_K)=\mathbb E_H[P_H(B_0)]\prod_{k=1}^{K}\bar P(B_k)=\prod_{k=0}^{K}\bar P(B_k)$. The $K{+}1$ readings are therefore i.i.d., and the proof above applies. Pooled statistics then have their null expectations, although the conditional rank is not uniform unless $P_H=\bar P$.

\paragraph{Decomposition of the expected CRPS.}
The continuous ranked probability score of a forecast CDF $F$ at an outcome $y$ is $\mathrm{CRPS}(F,y)=\int(F(x)-\mathbf 1\{y\le x\})^2\,dx$. For $Y\sim G$, $\mathbb E(F(x)-\mathbf 1\{Y\le x\})^2=(F(x)-G(x))^2+{G(x)(1-G(x))}$, giving
\begin{equation}
 \mathbb E\,\mathrm{CRPS}(F,Y)=\int G(1-G)\,dx+\int (F-G)^2\,dx .
 \label{eq:app-crps-decomp}
\end{equation}
Setting $F=Q_H$ or $\bar P$ with $G=P_H$ and averaging over $H$ yields expected CRPS $U+D_Q$ for the model and $U+I$ for climatology. This produces the skill formula \eqref{eq:skill}, with $D_Q=0$ only when $Q_H=P_H$. The fair ensemble estimator is unbiased given $H$ and $Y$~\cite{ferro2014fair}, and we estimate skill as a ratio of sums across histories.

\paragraph{Energy score.}
For $\mathrm{ES}(F,\mathbf y)=\mathbb E\|\mathbf X-\mathbf y\|-\tfrac12\mathbb E\|\mathbf X-\mathbf X'\|$, $\mathbb E\,\mathrm{ES}(F,\mathbf Y)-\mathbb E\,\mathrm{ES}(G,\mathbf Y)=\tfrac12\mathcal E(F,G)$, half the energy distance, which vanishes only when $F=G$~\cite{gneiting2007proper}. Whole trajectories therefore obey the same decomposition, with the fair estimator averaging over all $K(K{-}1)$ ordered pairs of distinct samples.

\paragraph{The climatological reference in practice.}
We approximate $\bar P$ using readings of the same quantity on histories of the development split (DEV) from the same family (raw real video), drawn from disjoint logs under the same past-only rules. For whole trajectories, the climatological ensemble consists of the family's DEV trajectories on the same time grid.

\subsection{The spread ratio}
\label{app:spread}
For one history, let $S=\mathbb E|X-X'|$ and $E=\mathbb E|X-Y|$, with $X,X'$ independent draws from $Q_H$. Under $\mathcal H_0$, $Y$ has the law of $X$ and is independent of it, so $E=S$ in every history and $\rho=\mathbb E_HS/\mathbb E_HE=1$. Given $K\ge2$ samples, $\hat S=\frac{1}{K(K-1)}\sum_{k\neq\ell}|X_k-X_\ell|$ and $\hat E=\frac1K\sum_k|X_k-Y|$ are unbiased for $S$ and $E$ given $H$, for every $K$. They are the two terms of the fair CRPS, $\hat E-\tfrac12\hat S$~\cite{ferro2014fair}. We estimate $\rho$ per family by the ratio of sums $\sum_i\hat S_i/\sum_i\hat E_i$ over histories and average over the scored families. The ratio of sums converges to $\mathbb E_HS/\mathbb E_HE$ for every $K$, so $\rho$, unlike the outside rate, does not depend on the number of samples. Samples that are too narrow or shifted give $E>S$ and $\rho<1$. For $P_H=\mathcal N(\mu,\sigma^2)$ and $Q_H=\mathcal N(\mu,w^2\sigma^2)$, $\rho=w\sqrt{2/(1+w^2)}$, so $\rho=0.76$ corresponds to samples whose spread is about 0.64 times the real one. The spread ratio was added after all results and is descriptive.

%% file: appendix/implementation.tex
\section{Implementation and reporting details}
\label{app:implementation}

\subsection{Generation, histories and instrument}
\input{tables/model_settings}
\label{app:generation}
\emph{Generation.} \Cref{tab:settings} lists the settings of each model. Models that accept text receive a prompt describing a forward-facing dashcam. With two exceptions, every model runs at its official native resolution and sampler settings, using distilled variants only where the developers release them. VaViM's released default decodes greedily, which would make all 16 samples identical, so we sample its full temperature-1 distribution. DrivingWorld's released inference reads the next recorded ego pose, so, as in its released long-term demo, every future step receives the last history pose step instead. No model receives any information from after $t_0$. The $k$-th sample of every model uses the same seed, derived from the history and $k$, and no sample is discarded or re-drawn. The 35{,}200 test futures (200 histories $\times$ 16 samples $\times$ 11 models) were generated on H100 and H200 GPUs, except those of Wan2.1-VACE, which were generated on RTX 6000 Ada GPUs.

\emph{Histories.} The family rules read only annotations and poses up to $t_0$. We verified this by hiding all information after each anchor. A family's target must be visible at $t_0$, be at least 20 pixels tall, carry at least five LiDAR points and be matched by a detection of the right class. The 200 test histories come from 200 logs, none of which appears in DEV, so test logs never enter instrument development.

\emph{Instrument.} The target of a family is fixed at $t_0$ and tracked through each continuation by an ensemble of three trackers on per-frame detections (median of the three). Its forward distance and lateral offset come from the ground-contact point of the tracked box under a flat-road model. Visual odometry on the same band gives the ego motion. The same code, parameters and failure rules are applied to every video.

\begin{figure*}[t]
\centering
\includegraphics[width=\linewidth]{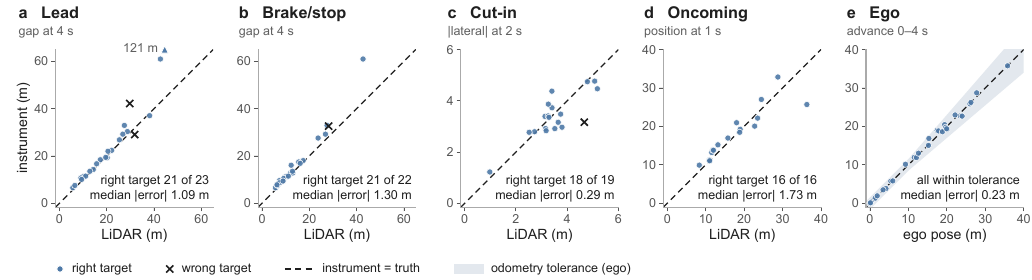}
\caption{\textbf{Instrument validation against ground truth on DEV}, shown on Orbis~2's reference clips: LiDAR annotations for the four target families (a--d, with 23, 22, 19 and 16 histories) and AV2 ego poses for the ego advance (e, 32 histories). Dashed line: exact agreement. Dots mark right-target readings, and crosses mark wrong-target readings counted by the reliability check. Shaded band in (e): odometry tolerance $\max(0.15\times\text{truth},1.5\,\text{m})$, containing all 32 readings. Each panel reports the right-target share and median absolute error. One lead reading lies above the plotted range in (a): 121\,m vs.\ 44.5\,m (triangle).}
\label{fig:validation}
\end{figure*}

\emph{Validation.} The check in \cref{tab:families} uses the raw real video of DEV. \Cref{fig:validation} repeats it on Orbis~2's reference clips, the real video reconstructed by Orbis~2's autoencoder, so its numbers differ slightly. On the raw video, the lead gap at 4\,s has a 90th-percentile error of 6.0\,m, and the ego advance over 0--4\,s a median error of 0.34\,m and a rank correlation of 0.997 with the AV2 poses. A model's ego family is scored only if its reference clips pass the odometry gate, which requires a rank correlation of at least 0.90 with the AV2 poses at 4\,s and median errors within fixed tolerances. The 14 secondary quantities, those tested for at least one model, include time headway, post-encroachment time, closing speed, lateral speed, the time an oncoming vehicle leaves the view, ego speed and peak deceleration.

\begin{figure*}[t]
\centering
\includegraphics[width=\linewidth]{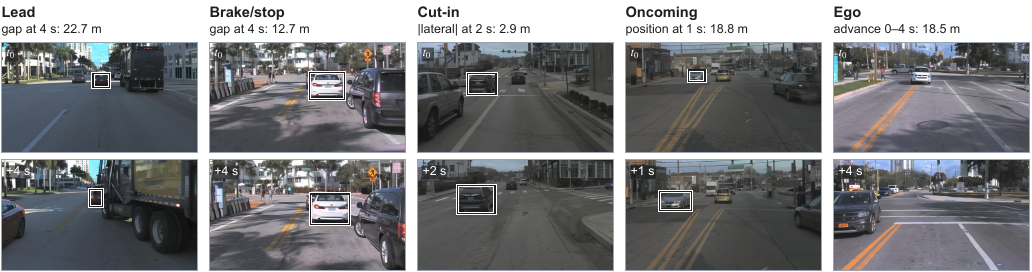}
\caption{\textbf{What each scenario family measures} (TEST, one typical history per family: the one whose real reading is closest to the family median). Each column gives the family's primary quantity and its real reading, with the frames at $t_0$ and at the quantity's horizon and the instrument's target box. The frames come from Orbis~2's reference clip, the real video after Orbis~2's autoencoder, which is what the instrument reads. The ego family reads the ego's own motion and has no target.}
\label{fig:families}
\end{figure*}

\Cref{fig:families} illustrates what each scenario family measures.

\begin{figure*}[t]
\centering
\includegraphics[width=\linewidth]{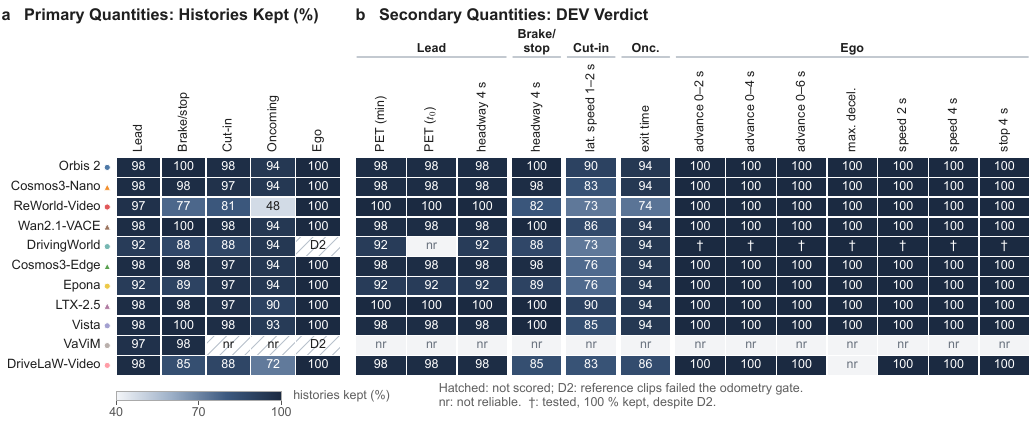}
\caption{\textbf{Measurable quantities per model.} Which quantities are tested was fixed on DEV before measuring any test clip; numbers are shares of TEST histories kept. \textbf{(a)}~Primary quantity of each family. Hatched cells are not scored because the model's reference clips failed the odometry gate (D2) or the reliability check (nr). \textbf{(b)}~DEV verdict for the secondary quantities: tested (shaded, showing retained share) or nr (not reliable: the readings disagree with the ground truth, or the raw and reconstructed real video give different readings). Quantities that almost no model passes are omitted. Eight are not reliable for at least nine of the eleven models, because the range noise of the box bottom dominates these derivatives of the gap: the closing speed at 2 and 4\,s, the lead speed at 4\,s and the minimum time to collision (TTC), in both Lead and Brake/stop (only Cosmos3-Nano and Cosmos3-Edge pass the brake/stop closing speed at 4\,s; \cref{app:details}). Six cannot be tested for at least ten models, because the event is too rare (hard braking in Lead and Brake/stop) or the reading is mostly censored (the post-encroachment times in Brake/stop, and the minimum TTC and entry time of a cut-in). $\dagger$: DrivingWorld's secondary ego quantities, read on all histories, passed their own validation with all histories kept, although its reference clips failed the primary odometry gate.}
\label{fig:scorability}
\end{figure*}

\subsection{Secondary statistics, scores and pre-ranks}
\label{app:secondary}
\emph{Secondary tests.} For each quantity we test the mean $\mathrm{PIT}-\tfrac12$, the convexity component of the rank chi-square (spread)~\cite{jolliffe2008rank} and the shares of cases below and above all samples (exact binomial), two-sided and with Benjamini--Hochberg control at $q=0.10$, and report a Monte Carlo chi-square test across the 17 rank bins.

\emph{Fair CRPS.} For $K$ samples $X_1,\dots,X_K$ and real outcome $Y$,
\begin{equation}
 \widehat{\mathrm{CRPS}}=\tfrac1K\sum_{k}|X_k-Y|-\tfrac{1}{2K(K-1)}\sum_{k\neq\ell}|X_k-X_\ell|
 \label{eq:fair-crps}
\end{equation}
is an unbiased estimator of the CRPS of the sampling distribution~\cite{ferro2014fair}. Whole trajectories use its multivariate counterpart, the fair energy score~\cite{gneiting2007proper}.

\emph{Pre-ranks.} Let $r_{jt}\in\{1,\dots,K{+}1\}$ be the rank of trajectory $j$ among the $K{+}1$ trajectories at time $t$, with ties broken at random. Band depth $\sum_t r_{jt}(K{+}2-r_{jt})$ measures how central a trajectory is, and the average rank $\mathrm{mean}_t\,r_{jt}$ orders trajectories from low to high, so that its two extremes play the role of H2 for whole trajectories.

\subsection{Records to be released}
We will release one record per reading.
Every table in the paper can be rebuilt from these records, and every exclusion keeps its original denominator.

\subsection{Retention and the scores}
The fair CRPS is unbiased for independent samples from the model.
Keeping a bundle only when all 17 readings are valid preserves the rank argument, but it can change the joint law of the samples. The fair CRPS and energy score therefore describe the predictive law given validity, not the unconditional law of the model.
When validity is decided for each video by the same rule, independent samples stay independent given their own validity, and both scores are unbiased for this conditioned law.

\subsection{Scope}
The benchmark tests the conditional predictions of the measured quantities in the futures that each model generated for it.
No model receives future ego actions (\cref{tab:settings}), so every test concerns open-loop forecasts from the history alone. The benchmark makes no claim about interventions, closed-loop control or methods that would improve the models' distributions.

%% file: tables/model_settings.tex
\begin{table*}[t]
\centering
\footnotesize
\setlength{\tabcolsep}{4pt}
\caption{\textbf{Generation settings} (from each model's frozen protocol). Frame: the size at which the model generates; every clip is then mapped onto the $512\times288$ measurement canvas and read at the 10\,Hz ticks (nearest frame). History: the frames of the 2\,s history that the model reads. Orbis~2's two sets of five frames share $t_0$. Guidance: classifier-free guidance scale; Vista's guider rises from 1.0 at the ends of each window to 2.5 at its middle, and ReWorld-Video uses its own internal guidance. Prompt: a text prompt describing a forward-facing dashcam. Calls: model calls per 6\,s future, or autoregressive steps.}
\label{tab:settings}
\begin{tabular}{@{}l l l l l l l l@{}}
\toprule
Model & Frame & Rate & History read & Other input & Sampler & Guidance & Calls \\
\midrule
Orbis~2 & $512\times288$ & 10\,Hz & 9 frames (5 at 10\,Hz, 5 at 2\,Hz) & -- & flow matching, NFE 5 and 20 & none & 12 steps \\
Cosmos3-Nano & $320\times192$ & 10\,Hz & 21 frames & prompt & UniPC, 35 steps & 6.0 & 1 \\
Cosmos3-Edge & $320\times192$ & 10\,Hz & 21 frames & prompt & UniPC, 35 steps & 6.0 & 1 \\
LTX-2.5 (distilled) & $1024\times576$ & 20\,Hz & 41 frames & prompt & Euler ancestral, 8 + 3 steps & none & 1 \\
Wan2.1-VACE (1.3B) & $832\times480$ & 16\,Hz & 33 frames & prompt & UniPC, 50 steps & 5.0 & 2 \\
Epona & $1024\times512$ & 5\,Hz & 10 frames (1.8\,s) & past ego poses & Euler, 100 steps & none & 30 steps \\
DrivingWorld & $512\times256$ & 10\,Hz & 15 frames & past ego poses & tokens, top-$k$ 30, $T{=}1$ & none & 60 steps \\
ReWorld-Video & $1024\times512$ & 8\,Hz & 9 frames (1.0\,s) & -- & flow matching, 30 steps & internal 1.4 & 2 \\
DriveLaW-Video & $1280\times704$ & 8\,Hz & 9 frames (1.0\,s) & prompt & 30 steps & 3.0 & 2 \\
VaViM & $512\times288$ & 2\,Hz & 5 frames & -- & tokens, full vocabulary, $T{=}1$ & none & 12 steps \\
Vista & $1024\times576$ & 10\,Hz & last frame & -- & Euler EDM, 50 steps & 1.0--2.5 & 3 \\
\bottomrule
\end{tabular}
\end{table*}

%% file: appendix/detailed_results.tex
\section{Per-model numbers behind the findings}
\label{app:details}
\renewcommand{\dblfloatpagefraction}{0.85}
\Cref{sec:experiments} states each finding with selected numbers, and this section gives all per-model values under matching headings.

\begin{figure*}[tp]
\centering
\includegraphics[width=\linewidth]{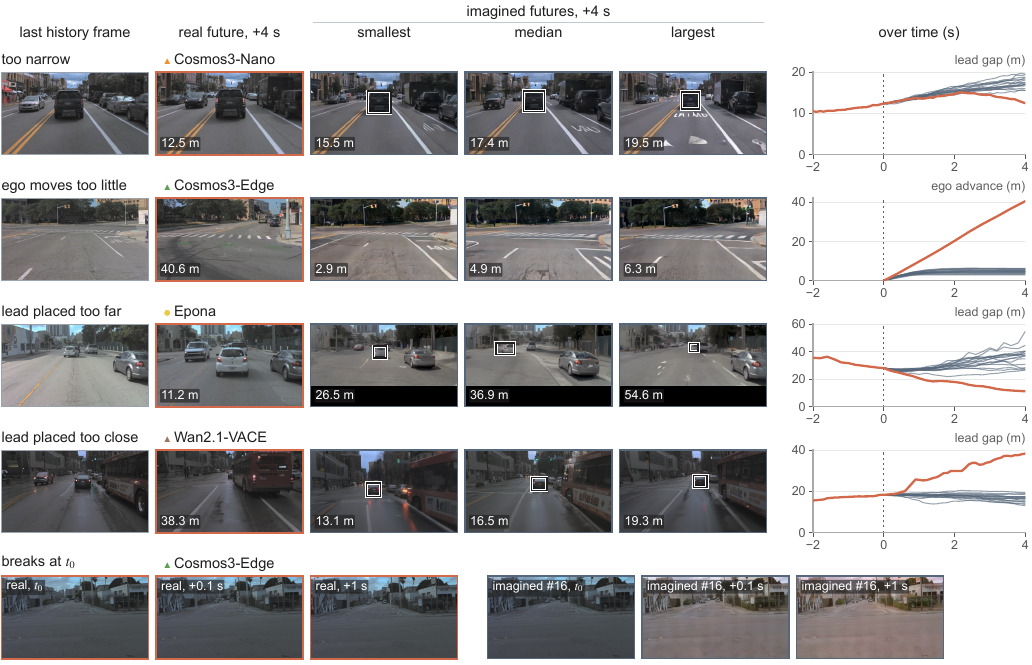}
\caption{\textbf{Failure modes} (TEST). Each row shows the most extreme test case of a model under a pre-set selection rule: samples that agree closely yet miss reality (Cosmos3-Nano), an ego vehicle that barely moves (Cosmos3-Edge), a lead vehicle placed too far ahead (Epona) or too close (Wan2.1-VACE), and a continuation that jumps at $t_0$ (Cosmos3-Edge, whose first step is 47.6 times larger than the history's steps, against a threshold of 3). Frames show the last history frame, the real future, and the imagined futures with the smallest, median and largest reading at $+4$\,s; the last row shows the real video and one imagined future at $t_0$, $+0.1$ and $+1$\,s. Plots track the quantity over time for imagined futures (slate) and reality (terracotta). Real frames show raw video, while readings are taken after each model's autoencoder.}
\label{fig:failures}
\end{figure*}

\begin{figure*}[tp]
\centering
\includegraphics[width=\linewidth]{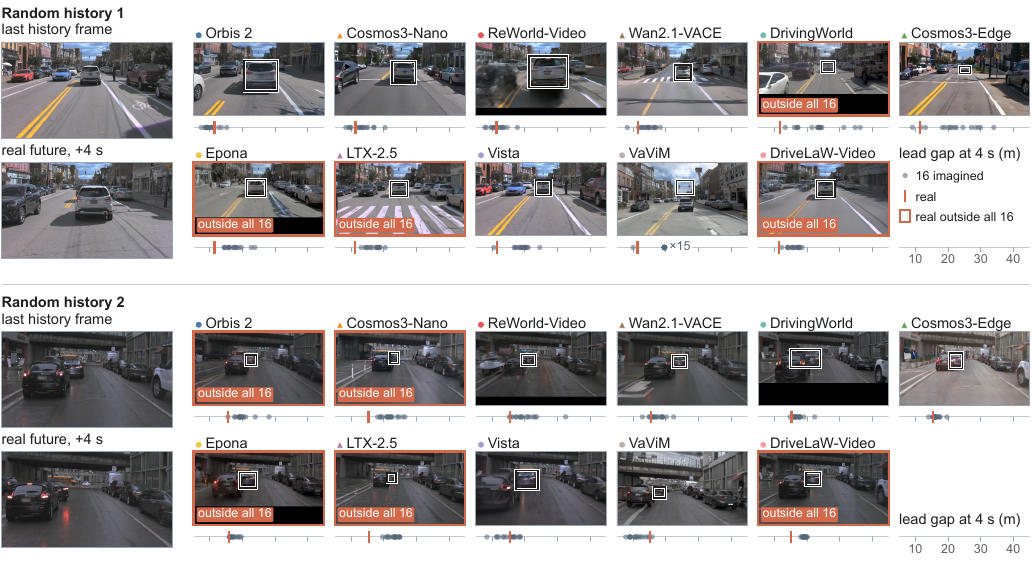}
\caption{\textbf{Two test histories drawn at random}, eleven models, in the layout of \cref{fig:gallery}. The histories were drawn with seed 2027 from the 28 lead-family test histories, among the 57 that every model scores, that have the lead followed at 4\,s and are not shown in another figure. Each model: its median imagined future at $+4$\,s with the instrument's box, and below it the 16 imagined lead gaps (slate) and the real gap (terracotta). Framed: the real gap lies outside all 16, for four models in history 1 and five in history 2. $\times15$: fifteen of VaViM's samples give the same reading.}
\label{fig:gallery-random}
\end{figure*}

\Cref{fig:failures} illustrates failure modes on extreme test cases, and \cref{fig:gallery-random} shows two randomly drawn test histories in the same layout as \cref{fig:gallery}. The typical history of that figure was chosen by a fixed rule. Among the 57 lead-family test histories that all eleven models score, the median number of models whose real gap lies outside all 16 samples is four, and ten histories have exactly four. In four of these ten the lead is tracked at 4\,s in the shown sample of every model, and one of the four was drawn with seed 2026.

\paragraph{Validity.}
Between 95.0\% and 98.3\% of the (history, family) cases are scorable for Orbis~2, the two Cosmos3 models, Epona, LTX-2.5, Wan2.1-VACE and Vista.
ReWorld-Video and DriveLaW-Video keep 82.2\% and 89.7\%, because their samples re-decode the last second of the history and a case is dropped when any sample loses the target at $t_0$. Dropping only those samples gives the same decisions (H2 28.9\% and 61.1\%).
DrivingWorld keeps 65.2\%, because its ego family is not scored, and VaViM keeps 35.4\%, because only its lead and brake/stop families are scored (97.7\% of their cases). On VaViM's reconstructed real video, the cut-in and oncoming readings miss the 90\% right-target bar (89.5\% and 68.8\%), and its odometry misses the 0.90 rank-correlation bar of the odometry gate (0.73), also when read at its own 2\,Hz rate. \Cref{fig:scorability} maps, per model, which quantities are tested.
Measured on its own 2\,Hz step, 1.3\% of VaViM's clips jump at $t_0$.

\begin{figure*}[tp]
\centering
\includegraphics[width=\linewidth]{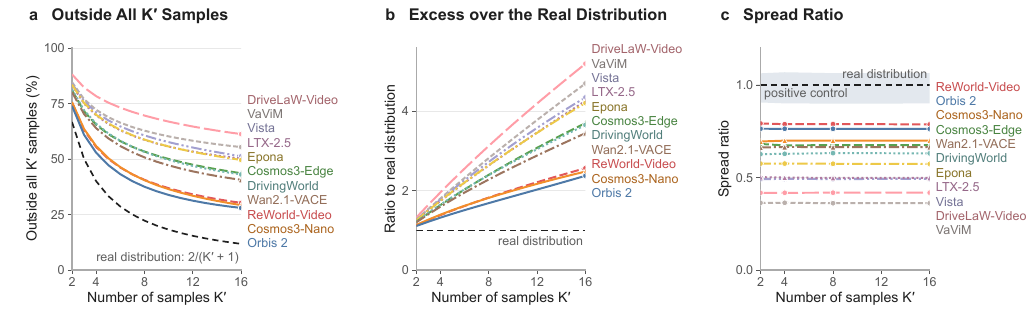}
\caption{\textbf{Underestimated uncertainty is not an artefact of $K=16$} (TEST). \textbf{(a)}~For $K'$ of the 16 samples drawn at random, the expected share of cases where reality falls outside all $K'$ (computed exactly over all subsets; history-averaged as in H2). Dashed: $2/(K'{+}1)$ under the real distribution. \textbf{(b)}~Ratio of that rate to its reference. A model that follows the real distribution stays at 1 for every $K'$, whereas every model's ratio grows with $K'$, from 1.11--1.32 at $K'=2$ to 2.38--5.20 at $K'=16$. \textbf{(c)}~The spread ratio does not change with $K'$ (\cref{tab:spread}). Shaded band: its positive control. Marker: model type ($\bullet$ driving, $\blacktriangle$ general).}
\label{fig:k-subsampling}
\end{figure*}

\paragraph{Coverage (H2).}
VaViM's 54.5\% is H2 over its two scored families, Lead and Brake/stop.
Across models, the range spanned by all 16 samples (nominally an 88.2\% prediction interval) covers only 38.8--72.0\% of real outcomes.
Two futures drawn at random from the 16 leave the real outcome outside all samples 1.11--1.32 times as often as expected under the real distribution, rising to 2.38--5.20 times for all 16 (\cref{fig:k-subsampling}). By contrast, the spread ratio stays within 0.01 of its 16-sample value at every $K'$, ranging from 0.36 (VaViM) to 0.79 (ReWorld-Video) with every 95\% interval below 1 (\cref{tab:spread}, \cref{app:spread}).
The outside rates of Orbis~2, Cosmos3-Nano and ReWorld-Video (28.0--30.2\%) lie closer to 11.8\% than those of the remaining eight models (40.6--61.2\%).
By family, Cosmos3-Edge, Epona, LTX-2.5, Vista and DriveLaW-Video miss the ego advance in 70--77\% of histories. The oncoming position is missed in 13.8--32.8\% of cases for every model that scores it except DriveLaW-Video (60.0\%), and it is the family closest to 11.8\% for seven of these ten models. Orbis~2 comes closest on the ego advance (23.8\%) and ReWorld-Video on the lead gap (20.6\%).

\begin{figure}[t]
\centering
\includegraphics[width=\linewidth]{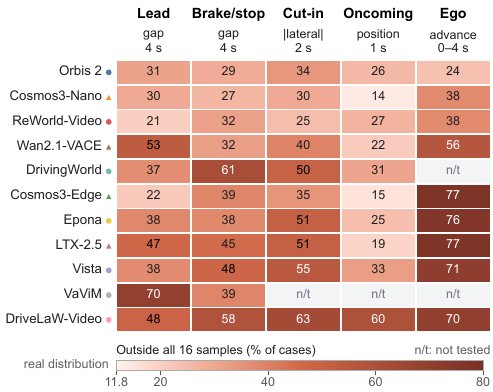}
\caption{\textbf{Outside rate by scenario family} (TEST). Share of cases (\%) where the real outcome falls outside all 16 samples, broken down by model and family. Under the real distribution this rate is 11.8\% (lightest shade). Rows follow the order of \cref{tab:main}. n/t: not tested.}
\label{fig:family-misses}
\end{figure}

\Cref{fig:family-misses} breaks this outside rate down by model and scenario family.

\input{tables/spread_ratio}

\paragraph{Skew (H1).}
Mean $\mathrm{PIT}-\tfrac12$ on the 4\,s lead gap for the five confirmed models is $-0.238$ ($z=-6.4$) for Epona, $-0.169$ ($z=-4.6$) for DrivingWorld, $+0.249$ ($z=+6.9$) for VaViM, $+0.177$ ($z=+4.9$) for Wan2.1-VACE and $+0.120$ ($z=+3.3$) for Vista, where $z$, the statistic of H1, is the slope component of the rank chi-square~\cite{jolliffe2008rank}, tested two-sided against the standard normal distribution.
Averaged across the scored primary quantities, the mean absolute offset spans from 4.1 (ReWorld-Video) and 4.2 (Orbis~2) up to 19.7 (Vista) on the $\times100$ scale.

\renewcommand{\dblfloatpagefraction}{0.9}
\input{tables/trajectory_families}
\begin{figure*}[t]
\centering
\includegraphics[width=\linewidth]{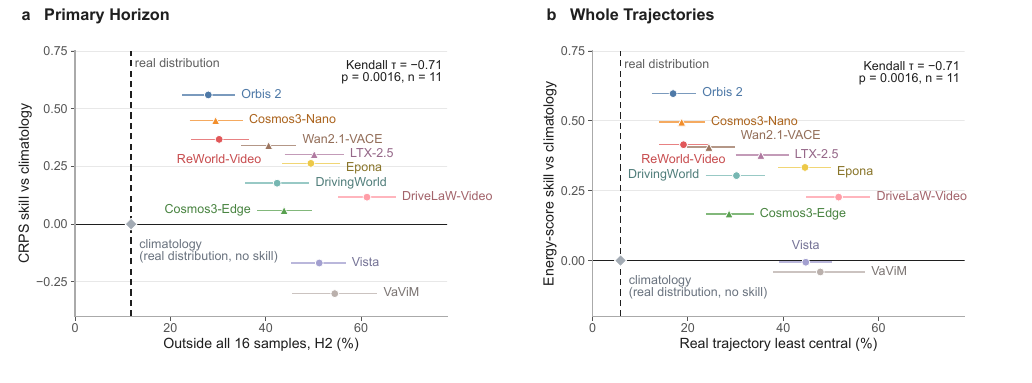}
\caption{\textbf{Informative samples still miss the real distribution} (TEST, exploratory). \textbf{(a)}~H2 (95\% intervals) vs.\ CRPS skill over climatology, averaged across scored families. \textbf{(b)}~Whole trajectories: least-central share vs.\ energy-score skill. Dashed lines mark the real distribution (11.8\%, 5.9\%). Climatology (diamond) lies on the dashed lines, because it passes pooled rank tests by construction, but has zero skill. Every model except Vista and VaViM has positive skill, and none follows the real distribution.}
\label{fig:skill-calibration}
\end{figure*}
\begin{figure*}[t]
\centering
\includegraphics[width=\linewidth]{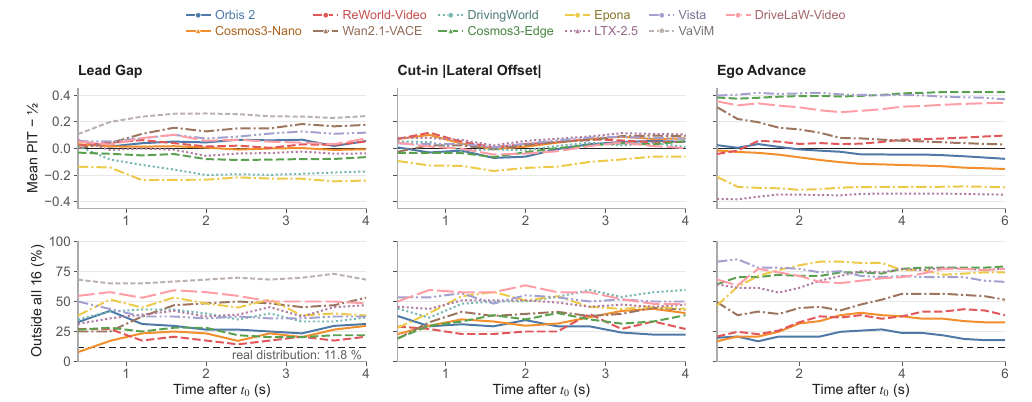}
\caption{\textbf{Skew and coverage along the trajectory} (TEST, exploratory). Top: mean $\mathrm{PIT}-\tfrac12$ of the real reading at each step after $t_0$ ($+$: generated values too small, $-$: too large). Bottom: share of cases where the real reading falls outside all 16 samples (dashed: 11.8\% under the real distribution). Columns show Lead gap, Cut-in $|$lateral offset$|$ and Ego advance on a 0.4\,s grid up to each quantity's validated horizon, measured with the same instrument, validity rule and cases as the primary quantities. Marker: model type ($\bullet$ driving, $\blacktriangle$ general), with legend above. Untested quantities have no line.}
\label{fig:horizon}
\end{figure*}

\paragraph{Whole trajectories and horizon.}
The real trajectory is least central among the 17 in 16.9--51.6\% of cases, or 2.9 to 8.8 times the 5.9\% expected under the real distribution (\cref{app:trajectory}).
For both the lead gap and the ego advance, every model except Cosmos3-Nano leaves the real value outside all samples in 20--83\% of cases within 0.4\,s of $t_0$ (\cref{fig:horizon}). Cosmos3-Nano starts near the real distribution and loses coverage over time (lead gap: 8\% at 0.4\,s vs.\ 30\% at 4\,s).

\paragraph{Shift and spread.}
Wan2.1-VACE lies between the U-shaped histogram of a model with too little spread and the one-sided histogram of a shifted model (16.4\% below and 24.2\% above all samples; \cref{fig:ranks}). The U-shaped histogram of DriveLaW-Video comes from opposite shifts: its ego moves too little, and its oncoming vehicles stay in view too long.

\paragraph{Ego motion.}
Headway and post-encroachment time follow the ego error. Cosmos3-Nano, Epona and LTX-2.5 move the generated ego too far, and all their significantly offset headways and post-encroachment times are too short. Cosmos3-Edge, Vista, DriveLaW-Video and DrivingWorld move it too little, and all of theirs are too long. ReWorld-Video and Wan2.1-VACE move it slightly too little and give post-encroachment times that are too long. The brake/stop closing speed passes the DEV reliability check only for Cosmos3-Nano ($-0.06$) and Cosmos3-Edge ($+0.13$, significant).
Along the horizon (\cref{fig:horizon}), ego offsets in Cosmos3-Edge, Epona, LTX-2.5, Vista and DriveLaW-Video reach full size within 0.8\,s and remain between 0.27 and 0.42 in magnitude through 6\,s.
Cosmos3-Nano and ReWorld-Video instead drift gradually, from $-0.01$ at 0.4\,s to $-0.15$ at 6\,s and from $-0.04$ to $+0.10$, respectively. DrivingWorld's lead-gap offset similarly builds from $+0.02$ at 0.4\,s to $-0.20$ at 2\,s.

\paragraph{Informativeness.}
Share of the climatological CRPS at the primary horizon and of the climatological energy score over whole trajectories that each model removes: Orbis~2 56\% and 60\%, Cosmos3-Nano 45\% and 50\%, ReWorld-Video 37\% and 41\%, Wan2.1-VACE 34\% and 41\%, LTX-2.5 30\% and 38\%, Epona 26\% and 33\%, DrivingWorld 18\% and 30\%, DriveLaW-Video 12\% and 23\%, Cosmos3-Edge 6\% and 17\%, Vista $-17\%$ and $-1\%$, and VaViM $-30\%$ and $-4\%$ over its two families.
The oncoming position reaches a CRPS skill of 0.42--0.77 for every model that scores it except Vista (0.19) and DriveLaW-Video (0.18).
Skill and the outside rate order the models alike (\cref{fig:skill-calibration}, Kendall $\tau=-0.71$, exact $p=0.0016$, $n=11$, for both the scalar and the trajectory pair), as \cref{eq:skill} predicts.
The single quantity closest to the real distribution is the time an oncoming vehicle leaves the view (LTX-2.5: mean $\mathrm{PIT}-\tfrac12=+0.001$, CRPS skill $+0.73$).

\paragraph{Robustness checks.}
Among the 51 scored model--family pairs, the largest positive-control offset occurs on DrivingWorld's brake/stop gap ($+0.085$, unadjusted $p=0.03$). Over whole trajectories, the positive control (16 generated trajectories, one as the observation) yields 3.8--9.8\% least-central cases (reference $1/16=6.25\%$) and 10.1--14.9\% outside cases (reference $2/16=12.5\%$).
In split-sample replication (the two halves of the 16 samples), Epona's H1 offset is $-0.224$ and $-0.225$ across the two halves, and VaViM's H1 slope $z$-score is $+6.6$ and $+6.4$.
Using raw real video as the observation gives H1 offsets of $-0.242$ (Epona), $-0.191$ (DrivingWorld), $+0.228$ (VaViM), $+0.182$ (Wan2.1-VACE) and $+0.112$ (Vista).
Treating lost targets as missing shifts Cosmos3-Edge's lead offset to $-0.075$ ($p=0.049$; its 95\% interval includes zero, so H1 stays unconfirmed) and increases VaViM's to $+0.339$.
The positive control for the spread ratio lies between 0.90 and 1.06 (\cref{tab:spread}).
Ties affect at most 8 of roughly 65 histories in any family.
Finally, Orbis~2 and LTX-2.5 include test futures from both H100 and H200 GPUs, and splitting by GPU type leaves the outside rate above the reference in both subsets.

%% file: tables/spread_ratio.tex
\begin{table*}[tp]
\centering
\footnotesize
\setlength{\tabcolsep}{5pt}
\caption{\textbf{The spread ratio does not depend on the number of samples} (TEST, descriptive). $\rho$: the mean distance between two samples over the mean distance from a sample to the real reading, averaged over the scored families, for $K'$ samples drawn at random from the 16 (200 draws per case). Under the real distribution, $\rho=1$ for every $K'$. Brackets: 95\% intervals over histories. Control: one generated sample as the observation, against 15 others (reference 1). Excess: the outside rate minus its value $2/(K'+1)$ under the real distribution, in points (at $K'=16$ from the frozen H2). Unlike $\rho$, it changes with $K'$. Models in the order of \cref{tab:main}.}
\label{tab:spread}
\begin{tabular}{@{}l cccc c cc@{}}
\toprule
 & \multicolumn{4}{c}{$\rho$} & & \multicolumn{2}{c}{Excess (points)} \\
\cmidrule(lr){2-5}\cmidrule(lr){7-8}
Model & $K'=2$ & $K'=4$ & $K'=8$ & $K'=16$ & Control & $K'=2$ & $K'=16$ \\
\midrule
Orbis 2 & 0.76 & 0.76 & 0.76 & 0.76\,{\scriptsize[0.69, 0.84]} & 0.97 & 7.3 & 16.2 \\
Cosmos3-Nano & 0.69 & 0.70 & 0.70 & 0.70\,{\scriptsize[0.63, 0.78]} & 0.99 & 9.3 & 17.7 \\
ReWorld-Video & 0.79 & 0.79 & 0.79 & 0.79\,{\scriptsize[0.72, 0.87]} & 0.90 & 8.7 & 18.5 \\
Wan2.1-VACE & 0.66 & 0.67 & 0.66 & 0.66\,{\scriptsize[0.61, 0.73]} & 1.01 & 13.4 & 28.8 \\
DrivingWorld & 0.63 & 0.63 & 0.63 & 0.63\,{\scriptsize[0.56, 0.71]} & 0.96 & 14.9 & 30.6 \\
Cosmos3-Edge & 0.68 & 0.67 & 0.68 & 0.67\,{\scriptsize[0.63, 0.74]} & 1.02 & 13.9 & 32.1 \\
Epona & 0.57 & 0.58 & 0.57 & 0.57\,{\scriptsize[0.52, 0.64]} & 1.06 & 16.9 & 37.7 \\
LTX-2.5 & 0.50 & 0.50 & 0.50 & 0.50\,{\scriptsize[0.44, 0.57]} & 0.97 & 16.6 & 38.4 \\
Vista & 0.49 & 0.49 & 0.49 & 0.49\,{\scriptsize[0.45, 0.54]} & 1.02 & 17.4 & 39.4 \\
VaViM & 0.36 & 0.36 & 0.36 & 0.36\,{\scriptsize[0.29, 0.45]} & 1.04 & 17.6 & 42.7 \\
DriveLaW-Video & 0.42 & 0.42 & 0.42 & 0.42\,{\scriptsize[0.37, 0.47]} & 1.02 & 21.6 & 49.5 \\
\bottomrule
\end{tabular}
\end{table*}

%% file: tables/trajectory_families.tex
\begin{table*}[p]
\centering
\footnotesize
\renewcommand{\arraystretch}{0.95}
\setlength{\tabcolsep}{4pt}
\caption{\textbf{Whole-trajectory results per family} (TEST, exploratory). LC: share of cases in which the real trajectory is the least central of the 17 by band depth (5.9\% under the real distribution). Out: share in which its average rank is the lowest or highest (11.8\% under the real distribution). $^\ast$: two-sided exact binomial test, BH $q\le0.10$ per column across models and families. Informativeness: energy-score skill against climatology (DEV trajectories of the same family). ES vs CV: energy-score skill against constant-velocity extrapolation (oncoming: no reliable history fit). Brackets: 95\% bootstrap intervals over histories.}
\label{tab:trajectory}
\begin{tabular}{@{}l l r r r l l@{}}
\toprule
Model & Trajectory & $n$ & LC\,\% & Out\,\% & Informativeness & ES vs CV \\
\midrule
Orbis 2 & Lead gap & 64 & 14.1$^\ast$ & 23.4$^\ast$ & $+0.55$ {\scriptsize[$+0.46$, $+0.63$]} & $+0.56$ {\scriptsize[$+0.45$, $+0.64$]} \\
 & Brake/stop gap & 65 & 20.0$^\ast$ & 27.7$^\ast$ & $+0.65$ {\scriptsize[$+0.57$, $+0.72$]} & $+0.75$ {\scriptsize[$+0.66$, $+0.81$]} \\
 & Cut-in lat. & 58 & 24.1$^\ast$ & 20.7$^\ast$ & $+0.39$ {\scriptsize[$+0.21$, $+0.55$]} & $+0.60$ {\scriptsize[$+0.43$, $+0.74$]} \\
 & Oncoming & 65 & 15.4$^\ast$ & 21.5$^\ast$ & $+0.79$ {\scriptsize[$+0.72$, $+0.84$]} & -- \\
 & Ego advance & 101 & 15.8$^\ast$ & 22.8$^\ast$ & $+0.61$ {\scriptsize[$+0.52$, $+0.69$]} & $+0.72$ {\scriptsize[$+0.65$, $+0.77$]} \\
\midrule
Cosmos3-Nano & Lead gap & 64 & 20.3$^\ast$ & 23.4$^\ast$ & $+0.50$ {\scriptsize[$+0.37$, $+0.64$]} & $+0.50$ {\scriptsize[$+0.35$, $+0.65$]} \\
 & Brake/stop gap & 64 & 20.3$^\ast$ & 17.2 & $+0.61$ {\scriptsize[$+0.45$, $+0.71$]} & $+0.72$ {\scriptsize[$+0.61$, $+0.81$]} \\
 & Cut-in lat. & 57 & 29.8$^\ast$ & 31.6$^\ast$ & $+0.26$ {\scriptsize[$+0.10$, $+0.42$]} & $+0.48$ {\scriptsize[$+0.27$, $+0.66$]} \\
 & Oncoming & 65 & 7.7 & 18.5 & $+0.77$ {\scriptsize[$+0.72$, $+0.82$]} & -- \\
 & Ego advance & 101 & 18.8$^\ast$ & 28.7$^\ast$ & $+0.34$ {\scriptsize[$+0.14$, $+0.51$]} & $+0.53$ {\scriptsize[$+0.44$, $+0.61$]} \\
\midrule
ReWorld-Video & Lead gap & 63 & 9.5 & 17.5 & $+0.46$ {\scriptsize[$+0.37$, $+0.56$]} & $+0.46$ {\scriptsize[$+0.33$, $+0.58$]} \\
 & Brake/stop gap & 50 & 26.0$^\ast$ & 30.0$^\ast$ & $+0.47$ {\scriptsize[$+0.34$, $+0.57$]} & $+0.60$ {\scriptsize[$+0.49$, $+0.70$]} \\
 & Cut-in lat. & 48 & 12.5$^\ast$ & 14.6 & $+0.30$ {\scriptsize[$+0.11$, $+0.46$]} & $+0.38$ {\scriptsize[$+0.23$, $+0.53$]} \\
 & Oncoming & 33 & 15.2$^\ast$ & 27.3$^\ast$ & $+0.53$ {\scriptsize[$+0.40$, $+0.64$]} & -- \\
 & Ego advance & 101 & 21.8$^\ast$ & 30.7$^\ast$ & $+0.31$ {\scriptsize[$+0.19$, $+0.41$]} & $+0.54$ {\scriptsize[$+0.46$, $+0.61$]} \\
\midrule
Wan2.1-VACE & Lead gap & 64 & 32.8$^\ast$ & 42.2$^\ast$ & $+0.28$ {\scriptsize[$+0.16$, $+0.42$]} & $+0.32$ {\scriptsize[$+0.12$, $+0.51$]} \\
 & Brake/stop gap & 65 & 18.5$^\ast$ & 23.1$^\ast$ & $+0.54$ {\scriptsize[$+0.46$, $+0.62$]} & $+0.68$ {\scriptsize[$+0.56$, $+0.77$]} \\
 & Cut-in lat. & 58 & 34.5$^\ast$ & 34.5$^\ast$ & $+0.21$ {\scriptsize[$+0.09$, $+0.33$]} & $+0.46$ {\scriptsize[$+0.26$, $+0.61$]} \\
 & Oncoming & 65 & 13.8$^\ast$ & 20.0$^\ast$ & $+0.73$ {\scriptsize[$+0.65$, $+0.79$]} & -- \\
 & Ego advance & 101 & 29.7$^\ast$ & 34.7$^\ast$ & $+0.27$ {\scriptsize[$+0.13$, $+0.38$]} & $+0.47$ {\scriptsize[$+0.37$, $+0.54$]} \\
\midrule
DrivingWorld & Lead gap & 60 & 28.3$^\ast$ & 33.3$^\ast$ & $+0.28$ {\scriptsize[$+0.14$, $+0.40$]} & $+0.34$ {\scriptsize[$+0.04$, $+0.55$]} \\
 & Brake/stop gap & 57 & 45.6$^\ast$ & 45.6$^\ast$ & $+0.25$ {\scriptsize[$+0.06$, $+0.43$]} & $+0.53$ {\scriptsize[$+0.36$, $+0.68$]} \\
 & Cut-in lat. & 52 & 44.2$^\ast$ & 40.4$^\ast$ & $-0.06$ {\scriptsize[$-0.33$, $+0.20$]} & $+0.33$ {\scriptsize[$+0.12$, $+0.48$]} \\
 & Oncoming & 65 & 13.8$^\ast$ & 20.0$^\ast$ & $+0.74$ {\scriptsize[$+0.67$, $+0.80$]} & -- \\
\midrule
Cosmos3-Edge & Lead gap & 64 & 15.6$^\ast$ & 26.6$^\ast$ & $+0.29$ {\scriptsize[$+0.15$, $+0.40$]} & $+0.29$ {\scriptsize[$+0.10$, $+0.44$]} \\
 & Brake/stop gap & 64 & 29.7$^\ast$ & 42.2$^\ast$ & $+0.38$ {\scriptsize[$+0.21$, $+0.53$]} & $+0.56$ {\scriptsize[$+0.43$, $+0.66$]} \\
 & Cut-in lat. & 57 & 24.6$^\ast$ & 33.3$^\ast$ & $+0.17$ {\scriptsize[$+0.04$, $+0.28$]} & $+0.41$ {\scriptsize[$+0.22$, $+0.56$]} \\
 & Oncoming & 65 & 6.2 & 13.8 & $+0.61$ {\scriptsize[$+0.45$, $+0.74$]} & -- \\
 & Ego advance & 101 & 53.5$^\ast$ & 70.3$^\ast$ & $-0.62$ {\scriptsize[$-0.82$, $-0.41$]} & $-0.15$ {\scriptsize[$-0.47$, $+0.10$]} \\
\midrule
Epona & Lead gap & 60 & 38.3$^\ast$ & 45.0$^\ast$ & $+0.30$ {\scriptsize[$+0.07$, $+0.46$]} & $+0.28$ {\scriptsize[$-0.01$, $+0.47$]} \\
 & Brake/stop gap & 58 & 34.5$^\ast$ & 41.4$^\ast$ & $+0.42$ {\scriptsize[$+0.29$, $+0.57$]} & $+0.64$ {\scriptsize[$+0.52$, $+0.73$]} \\
 & Cut-in lat. & 57 & 42.1$^\ast$ & 36.8$^\ast$ & $+0.19$ {\scriptsize[$-0.08$, $+0.40$]} & $+0.50$ {\scriptsize[$+0.26$, $+0.66$]} \\
 & Oncoming & 65 & 20.0$^\ast$ & 32.3$^\ast$ & $+0.75$ {\scriptsize[$+0.66$, $+0.81$]} & -- \\
 & Ego advance & 101 & 71.3$^\ast$ & 73.3$^\ast$ & $+0.01$ {\scriptsize[$-0.23$, $+0.21$]} & $+0.27$ {\scriptsize[$+0.13$, $+0.40$]} \\
\midrule
LTX-2.5 & Lead gap & 64 & 34.4$^\ast$ & 29.7$^\ast$ & $+0.44$ {\scriptsize[$+0.35$, $+0.55$]} & $+0.47$ {\scriptsize[$+0.32$, $+0.62$]} \\
 & Brake/stop gap & 64 & 29.7$^\ast$ & 42.2$^\ast$ & $+0.48$ {\scriptsize[$+0.36$, $+0.58$]} & $+0.64$ {\scriptsize[$+0.49$, $+0.75$]} \\
 & Cut-in lat. & 57 & 33.3$^\ast$ & 29.8$^\ast$ & $+0.25$ {\scriptsize[$+0.10$, $+0.41$]} & $+0.48$ {\scriptsize[$+0.28$, $+0.65$]} \\
 & Oncoming & 62 & 17.7$^\ast$ & 27.4$^\ast$ & $+0.79$ {\scriptsize[$+0.74$, $+0.84$]} & -- \\
 & Ego advance & 101 & 53.5$^\ast$ & 65.3$^\ast$ & $-0.08$ {\scriptsize[$-0.36$, $+0.14$]} & $+0.21$ {\scriptsize[$+0.10$, $+0.32$]} \\
\midrule
Vista & Lead gap & 64 & 26.6$^\ast$ & 31.2$^\ast$ & $+0.16$ {\scriptsize[$+0.03$, $+0.31$]} & $+0.23$ {\scriptsize[$-0.02$, $+0.46$]} \\
 & Brake/stop gap & 65 & 41.5$^\ast$ & 55.4$^\ast$ & $+0.06$ {\scriptsize[$-0.22$, $+0.27$]} & $+0.35$ {\scriptsize[$+0.13$, $+0.53$]} \\
 & Cut-in lat. & 58 & 37.9$^\ast$ & 44.8$^\ast$ & $+0.02$ {\scriptsize[$-0.13$, $+0.15$]} & $+0.32$ {\scriptsize[$+0.16$, $+0.44$]} \\
 & Oncoming & 64 & 25.0$^\ast$ & 31.2$^\ast$ & $+0.43$ {\scriptsize[$+0.26$, $+0.57$]} & -- \\
 & Ego advance & 101 & 69.3$^\ast$ & 73.3$^\ast$ & $-0.70$ {\scriptsize[$-0.90$, $-0.50$]} & $-0.23$ {\scriptsize[$-0.58$, $+0.02$]} \\
\midrule
VaViM & Lead gap & 63 & 66.7$^\ast$ & 69.8$^\ast$ & $-0.17$ {\scriptsize[$-0.36$, $-0.01$]} & $-0.02$ {\scriptsize[$-0.44$, $+0.32$]} \\
 & Brake/stop gap & 64 & 29.7$^\ast$ & 43.8$^\ast$ & $+0.09$ {\scriptsize[$-0.15$, $+0.28$]} & $+0.42$ {\scriptsize[$+0.15$, $+0.61$]} \\
\midrule
DriveLaW-Video & Lead gap & 64 & 42.2$^\ast$ & 43.8$^\ast$ & $+0.35$ {\scriptsize[$+0.23$, $+0.46$]} & $+0.34$ {\scriptsize[$+0.18$, $+0.49$]} \\
 & Brake/stop gap & 55 & 56.4$^\ast$ & 60.0$^\ast$ & $+0.45$ {\scriptsize[$+0.34$, $+0.56$]} & $+0.56$ {\scriptsize[$+0.42$, $+0.67$]} \\
 & Cut-in lat. & 52 & 44.2$^\ast$ & 38.5$^\ast$ & $+0.19$ {\scriptsize[$+0.02$, $+0.34$]} & $+0.31$ {\scriptsize[$+0.18$, $+0.45$]} \\
 & Oncoming & 50 & 60.0$^\ast$ & 72.0$^\ast$ & $+0.44$ {\scriptsize[$+0.27$, $+0.56$]} & -- \\
 & Ego advance & 101 & 57.4$^\ast$ & 67.3$^\ast$ & $-0.29$ {\scriptsize[$-0.48$, $-0.12$]} & $+0.17$ {\scriptsize[$-0.06$, $+0.35$]} \\
\bottomrule
\end{tabular}
\end{table*}

%% file: appendix/trajectory_results.tex
\section{Whole-trajectory results}
\label{app:trajectory}
\Cref{tab:trajectory} reports per-family whole-trajectory statistics corresponding to \cref{sec:whole-trajectories}. Trajectories are sampled on a 0.4\,s grid up to each quantity's validated horizon: 4\,s for Lead gap, Brake/stop gap and Cut-in lateral offset, 1\,s for Oncoming position (on a 0.2\,s grid), and 6\,s for Ego advance.

%% file: appendix/property_scores.tex
\section{Property scores}
\label{app:scores}
\Cref{fig:overview} summarizes each model by four property scores in $[0,1]$, with 1 ideal, and by their mean. The scores were defined after the frozen analyses and weight the properties equally. They summarize the test results and are not tests.
\begin{itemize}
\item \textbf{Validity} $S_{\mathrm{val}}=\mathrm{Scor}\,(1-J)$, the share of usable continuations. $\mathrm{Scor}$ is the share of (history, family) cases over all five families in which the primary quantity's real reading, taken after the model's autoencoder, and all 16 generated readings are valid. $J$ is the share of generated clips that break at $t_0$. A clip breaks when its first generated step changes the frame by more than three times the clip's mean frame-to-frame change over the last second of the history, measured on greyscale frames (reference clips: 0--1.5\%). A family that cannot be read on the model's own reconstruction of real video (DrivingWorld's ego family and VaViM's cut-in, oncoming and ego families) counts as unusable, since none of the model's futures can then be read for it. VaViM repeats the $t_0$ frame until its first 2\,Hz frame, so for VaViM $J$ compares that frame with $t_0$ against its own 0.5\,s history steps.
\item \textbf{Coverage} $S_{\mathrm{cov}}=\tfrac12[c(T_{\mathrm{H2}},\tfrac{2}{17})+c(T_{\mathrm{LC}},\tfrac{1}{17})]$ with $c(T,p)=1-|T-p|/(1-p)$. $T_{\mathrm{H2}}$ is the confirmatory H2 statistic (history-averaged share of cases in which the real outcome lies outside all 16 samples, $2/17$ under the real distribution), and $T_{\mathrm{LC}}$ is the same history-averaged share for the event that the real trajectory is the least central of the 17 by band depth ($1/17$ under the real distribution). For a model that underestimates its uncertainty ($T\ge p$), $c$ is the covered share relative to the covered share under the real distribution, $(1-T)/(1-p)$. For samples that are too wide ($T<p$) it decreases as well, so both directions of error lower the score.
\input{tables/property_scores}
\item \textbf{Centering} $S_{\mathrm{cen}}=1-2\,\mathrm{mean}_f|\overline{\mathrm{PIT}}_f-\tfrac12|$, the mean over the primary quantities of the model's scored families (five for most models). It is 1 when no family has a mean offset and 0 when every family has the largest possible offset, $\tfrac12$.
\item \textbf{Informativeness} $S_{\mathrm{inf}}=\min\{1,\allowbreak\max\{0,\allowbreak\mathrm{mean}_f\tfrac12(\mathrm{CRPSS}_f+\mathrm{ESS}_f)\}\}$, the average over the model's scored families of the CRPS skill of the primary quantity and the energy-score skill of its whole trajectory, both against climatology (the share of the climatological error that the model removes). Climatology scores 0 and the true conditional law scores the largest attainable value, which is below 1 whenever outcomes remain uncertain given the history.
\item \textbf{Summary} $S=\tfrac14(S_{\mathrm{val}}+S_{\mathrm{cov}}+S_{\mathrm{cen}}+S_{\mathrm{inf}})$.
\end{itemize}
\emph{Worked example (Orbis~2).} $\mathrm{Scor}=353/359=0.983$ and $J=0$, so $S_{\mathrm{val}}=0.983$. $T_{\mathrm{H2}}=0.280$ and $T_{\mathrm{LC}}=0.169$ give $c=0.720/0.882=0.817$ and $0.831/0.941=0.883$, so $S_{\mathrm{cov}}=0.850$. The mean absolute offset is $0.042$, so $S_{\mathrm{cen}}=0.916$. The CRPS and energy-score skills average $0.560$ and $0.598$, so $S_{\mathrm{inf}}=0.579$. Hence $S=(0.983+0.850+0.916+0.579)/4=0.832$. \Cref{tab:scores} lists all models.

%% file: tables/property_scores.tex
\begin{table}[t]
\centering
\footnotesize
\setlength{\tabcolsep}{3pt}
\caption{\textbf{Property scores} (TEST, definitions in \cref{app:scores}, 1 is ideal). Rows sorted by the summary score.}
\label{tab:scores}
\begin{tabular}{@{}l l r r r r r@{}}
\toprule
Model & Type & $S_{\mathrm{val}}$ & $S_{\mathrm{cov}}$ & $S_{\mathrm{cen}}$ & $S_{\mathrm{inf}}$ & $S$ \\
\midrule
Orbis 2 & driving & 0.983 & 0.850 & 0.916 & 0.579 & 0.832 \\
Cosmos3-Nano & general & 0.957 & 0.832 & 0.868 & 0.472 & 0.782 \\
ReWorld-Video & driving & 0.819 & 0.825 & 0.919 & 0.391 & 0.738 \\
Wan2.1-VACE & general & 0.961 & 0.738 & 0.797 & 0.373 & 0.717 \\
LTX-2.5 & general & 0.969 & 0.626 & 0.779 & 0.339 & 0.678 \\
Epona & driving & 0.950 & 0.581 & 0.661 & 0.298 & 0.623 \\
DrivingWorld & driving & 0.652 & 0.697 & 0.766 & 0.241 & 0.589 \\
Cosmos3-Edge & general & 0.832 & 0.697 & 0.688 & 0.113 & 0.583 \\
DriveLaW-Video & driving & 0.894 & 0.477 & 0.638 & 0.172 & 0.545 \\
Vista & driving & 0.966 & 0.570 & 0.606 & 0.000 & 0.536 \\
VaViM & driving & 0.349 & 0.536 & 0.708 & 0.000 & 0.398 \\
\bottomrule
\end{tabular}
\end{table}

%% file: appendix/scalar_validation.tex
\section{Validation of the rank test on simulated numbers}
\label{app:scalar}
\input{tables/statistical_validation}
Before applying rank tests to video, we checked the randomized ranks and a total-variation uniformity test on simulated numbers, where the right answer is known.

Each synthetic dataset has $N$ observations, with $N\in\{64,256,1024\}$ spanning our test-pool sizes, and each observation is paired with $K=16$ model draws as in the benchmark. Real observations follow $\mathcal N(0,1)$, while model samples follow one of four distributions: $\mathcal N(0,1)$ (correct), $\mathcal N(0.5,1)$ (mean shifted by half a standard deviation), $\mathcal N(0,0.5^2)$ (too narrow) or $\mathcal N(0,2^2)$ (too wide). In each dataset we rank every observation among its 16 samples and test uniformity across the 17 rank bins using total-variation distance against 50{,}000 simulated uniform histograms at level $\alpha=0.05$. We repeat each setting 2{,}000 times, and \cref{tab:statistical_validation} reports the empirical rejection rates.

Video readings can tie, for example when a lost target keeps its last value. To test the handling of ties, we draw both observations and samples from $\mathrm{Bernoulli}(0.3)$, so the model is exact and most comparisons are ties. We compare our randomized tie-breaking rule against strict inequalities that count only samples strictly below the observation.

Under the correct Gaussian model, the rejection rates of 4.2--4.8\% match the nominal 5\% level. Shifted, overly narrow and overly wide models are rejected in 52--96\% of datasets at $N=64$ and in almost all datasets for $N\ge 256$. Under the exact Bernoulli model, randomized tie-breaking rejects in 4.5--5.9\% of datasets, whereas strict comparison pushes ties to the low end and falsely rejects the exact model in every dataset.

%% file: tables/statistical_validation.tex
\begin{table}[!b]
\centering
\small
\setlength{\tabcolsep}{3pt}
\caption{\textbf{How often the rank test rejects, on simulated numbers} (\% of 2{,}000 datasets, $K=16$, $\alpha=0.05$). Gaussian rows: observed values from $\mathcal N(0,1)$, model samples from $\mathcal N(0,1)$ (correct), $\mathcal N(0.5,1)$ (mean shifted), $\mathcal N(0,0.5^2)$ (too narrow) or $\mathcal N(0,2^2)$ (too wide). Bernoulli rows: both from $\mathrm{Bernoulli}(0.3)$, so the model is correct and most comparisons are ties. A correct model should be rejected about 5\% of the time. All 95\% Wilson intervals lie within 2.2 points of the value shown.}
\label{tab:statistical_validation}
\begin{tabular}{@{}l r r r@{}}
\toprule
Model samples & $N=64$ & $N=256$ & $N=1024$ \\
\midrule
Gaussian, correct & 4.5 & 4.2 & 4.8 \\
Gaussian, mean $0.5$ & 52.4 & 99.9 & 100.0 \\
Gaussian, $\sigma=0.5$ & 95.8 & 100.0 & 100.0 \\
Gaussian, $\sigma=2$ & 92.0 & 100.0 & 100.0 \\
Bernoulli, random tie-breaks & 4.5 & 4.5 & 5.9 \\
Bernoulli, strict comparisons & 100.0 & 100.0 & 100.0 \\
\bottomrule
\end{tabular}
\end{table}